\documentclass[letterpaper]{article}
\usepackage[preprint]{aaai2027}
\usepackage[hyphens]{url}
\usepackage{graphicx}
\usepackage[numbers,square]{natbib}
\usepackage{caption}
\makeatletter
\let\aaai@real@bibliographystyle\bibliographystyle
\renewcommand{\bibliographystyle}[1]{%
  \def\aaai@arg{#1}%
  \def\aaai@blocked{aaai2027}%
  \ifx\aaai@arg\aaai@blocked
  \else
    \aaai@real@bibliographystyle{#1}%
  \fi
}
\makeatother

\usepackage{amsmath}
\usepackage{amssymb}
\usepackage{amsfonts}

\usepackage{algorithm}
\usepackage{algorithmic}

\usepackage{booktabs}
\usepackage{multirow}
\usepackage{float}
\usepackage{tabularx}
\usepackage{colortbl}
\definecolor{dmmblue}{HTML}{505190} 
\usepackage[colorlinks=true,linkcolor=dmmblue,citecolor=dmmblue,urlcolor=dmmblue]{hyperref}
\makeatletter
\AddToHook{begindocument}[dmm/hide-hyperref]{%
  \expandafter\let\expandafter\dmm@verhyperref\csname ver@hyperref.sty\endcsname
  \expandafter\let\csname ver@hyperref.sty\endcsname\relax}
\AddToHook{begindocument}[dmm/show-hyperref]{%
  \expandafter\let\csname ver@hyperref.sty\endcsname\dmm@verhyperref}
\DeclareHookRule{begindocument}{dmm/hide-hyperref}{before}{aaai2027}
\DeclareHookRule{begindocument}{dmm/show-hyperref}{after}{aaai2027}
\makeatother
\newcommand{\appref}[1]{\ifdefined\hyperref\hyperref[#1]{Appendix~\ref*{#1}}\else Appendix~\ref{#1}\fi}
\definecolor{best}{HTML}{ccd2fc}

\title{Decentralized Master-Mind: Joint Action Refinement through \\Iterative Intent Denoising in Multi-Agent Pathfinding}

\author{
    Valeriy Vyaltsev, 
    Anton Andreychuk, 
    Taisia Zlotnikova, \\
    Konstantin Yakovlev, 
    Aleksandr Panov, 
    Alexey Skrynnik\footnote{Corresponding author: \texttt{skrynnikalexey@gmail.com}}
}
\affiliations{
    CogAI Lab, Moscow, Russia
}

\begin{document}

\maketitle

\begin{abstract}
Decentralized multi-agent path finding (MAPF) with communication requires
agents to reach individual goals without collisions under partial observability.
Learnable policies trained on expert data provide an effective approach to this
problem. However, when several coordinated joint actions are valid in the same
context, independently sampling from per-agent distributions can recombine
locally valid choices into incompatible joint actions. This failure can arise
from the final sampling mechanism even when the per-agent action distributions
are learned correctly.
\textbf{DMM} (Decentralized Master-Mind) addresses this by replacing one-shot
action sampling with discrete, iterative refinement of action intents across
communication rounds, inspired by denoising in diffusion models. Agents
initialize random action intents and refine them through local communication,
coupling their choices before commitment. DMM is pretrained with imitation
learning on expert MAPF solutions and further optimized with
\textbf{MICPO}, a critic-free group-relative reinforcement-learning method
designed for multi-agent, multi-round action refinement. DMM generally achieves higher success rates and lower solution costs than the
evaluated learnable baselines.
On 1,600 MovingAI tasks, DMM fine-tuned with MICPO solves 1,598, the highest
coverage among the evaluated methods, while achieving solution costs close to
those of the strongest baselines. DMM
also scales to \textbf{over one million} simultaneously acting agents in
obstacle-rich environments. These results show that round-level intent
refinement can improve joint-action coordination while
preserving decentralized execution.
\end{abstract}

\begin{links}
    \link{Code}{https://github.com/CognitiveAISystems/DMM}
\end{links}

\section{Introduction}

\begin{figure}[t!]
    \centering
    \includegraphics[width=1.0\linewidth]{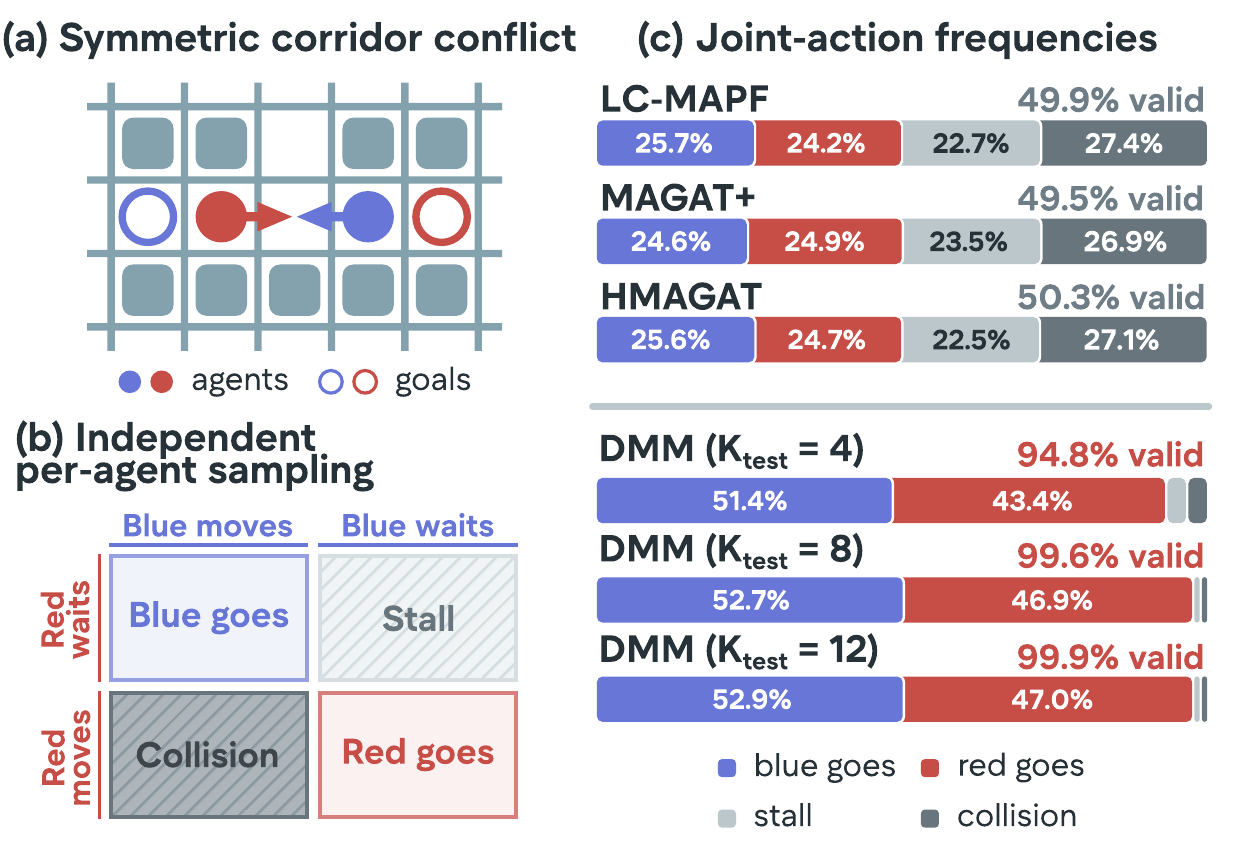}
    \caption{A symmetric corridor conflict illustrates the limitation of independent
per-agent action sampling and the effect of iterative intent refinement. All
policies are trained solely on this corridor scenario.
\textbf{(a)} Two agents start at opposite ends of a corridor and must swap
positions by reaching their assigned goals.
\textbf{(b)} At the shared ambiguous state, independent sampling permits four
joint outcomes: \emph{Blue goes} and \emph{Red goes} are valid, mutual waiting
causes a stall, and simultaneous movement causes a collision.
\textbf{(c)} Joint-action frequencies obtained from 1,000 samples per
independently trained checkpoint, averaged across five checkpoints. LC-MAPF, MAGAT+, and HMAGAT sample
each agent's action independently and distribute probability nearly uniformly
over the four outcomes, yielding about 50\% valid joint actions. DMM instead
concentrates probability on the two valid outcomes. With checkpoints trained at $K_{\mathrm{train}}=4$, increasing the test-time
refinement depth from $K_{\mathrm{test}}=4$ to $8$ and $12$ raises the
displayed valid joint-action frequency from 94.8\% to 99.6\% and
99.9\%, respectively.}
    \label{fig:corridor}
\end{figure}

Robotics, autonomous transportation, and logistics increasingly depend on large
teams of autonomous agents that must coordinate safely as their numbers grow,
from warehouse fleets to city-scale autonomous transport. Multi-Agent Path
Finding (MAPF) is one of the core problems in this domain~\cite{stern2019multi}:
a set of agents must navigate from their start locations to designated goal
vertices on a shared graph while avoiding collisions with one another. Despite
its seemingly simple formulation, MAPF is computationally challenging due to
the combinatorial explosion of joint configurations~\citep{surynek2010optimization}.
Classical approaches rely on centralized solvers that compute globally
optimal or near-optimal joint trajectories~\citep{sharon2015conflict,wagner2011m},
but their cost grows rapidly with the number of agents.

Decentralization offers an appealing alternative: agents act from bounded local
observations and execute in parallel, so per-agent computation need not grow
with team size. Deciding from partial information, however, comes at a cost to
solution quality. Learnable policies close much of this gap, allowing
coordination behavior to be acquired from data. This has been pursued through reinforcement learning~\citep{phan2024confidence,
skrynnik2024learn, skrynnik2024decentralized, phan2025generative} and imitation
learning~\citep{andreychuk2025mapf, veerapaneni2025work}, and further through
learned communication~\citep{ma2021distributed, ma2022learning,
li2021message, wang2023scrimp, jain2026pairwise}, which lets an agent condition
its decision on information from its neighbors.

Despite differences in architecture and communication, these policies
ultimately produce a separate action distribution for each agent and sample
their final actions separately. Whenever several joint actions are equally
valid in the same context, the joint-action distribution is multimodal, and a
policy that has learned to cover it assigns probability to multiple valid
modes. Independent draws can then recombine locally valid choices into
incompatible joint actions. For example, in a minimal corridor where two
agents must swap through a single-width passage, only two of the four combinations correspond to coordinated resolutions,
while the others produce a stall or a collision; independent sampling realizes
all four at roughly equal rates
(Figure~\ref{fig:corridor}). Communication can change each agent's action
distribution by enriching the information it conditions on, but once the
information available at action commitment is fixed, separate final sampling
still cannot represent residual dependence between the agents' choices.
Thus, even correctly learned local action distributions can produce
incompatible joint actions. What is missing is a way for agents to coordinate
their stochastic choices before commitment.

Sampling multimodal distributions is a central motivation for generative
models such as diffusion and flow matching, which use iterative refinement. We introduce \textbf{Decentralized Master-Mind
(DMM)}, which replaces one-shot action sampling with iterative refinement of
decentralized action intents. Each agent maintains an intent and refines it
over several communication rounds using information from neighboring agents,
so that decisions can influence one another before commitment. Unlike
standard generative refinement, which typically operates on a sample in
isolation, DMM makes the intermediate intent itself a part of the communication
process, preserving decentralized execution while coupling the agents'
decisions.

To improve solution quality, DMM is trained in two stages. Imitation on expert
MAPF trajectories teaches the policy to reproduce the expert's actions, but
matching actions does not directly optimize the quality of the resulting
joint solution. We therefore further optimize complete rollouts using
reinforcement learning. Standard actor-critic fine-tuning is challenging in
decentralized MAPF: a centralized critic must generalize over a combinatorial
joint state, while a decentralized critic has only partial information when
predicting a shared team outcome. We address this with
\textbf{Multi-agent Iterative Commitment Policy Optimization (MICPO)}, which
removes the critic and instead compares rollouts under matched conditions,
adapting group-relative optimization~\citep{deepseek-math} to the
multi-agent, multi-round structure of DMM.

Scale is where decentralization is supposed to pay off, and where coordination
through communication is hardest. We therefore evaluate DMM at three
complementary scales. On POGEMA~\citep{skrynnik2025pogema}, we compare DMM with other decentralized approaches using only the predictions of their learned policies, without downstream action correction, showing that DMM generally maintains higher success rates as team size grows
and achieves lower solution costs across the evaluated domains. On the MovingAI~\citep{stern2019multi} benchmark, we instead evaluate DMM with post-sampling
action processing as part of MAPF solvers across 1,600 large and diverse tasks, showing
that the resulting solver achieves the highest task success among the evaluated
methods, solving 1,598 of 1,600 tasks. Finally, GPU-resident execution and lightweight
inference adaptations enable DMM to solve all tested instances with obstacles and
over one million simultaneous agents.

\noindent Overall, the main contributions of this work are as follows:
\begin{itemize}
    \item We identify a structural limitation of learnable decentralized MAPF policies: even with communication, independently sampling each agent's final action can recombine individually valid choices into conflicting joint actions.
    \item We propose \textbf{DMM} (Decentralized Master-Mind), which reframes joint action selection as iterative refinement of action intents across communication rounds.
    \item We introduce \textbf{MICPO}, a critic-free multi-agent reinforcement learning method that fine-tunes DMM's multi-round refinement on trajectory-level outcomes.
    \item We demonstrate the effectiveness of DMM in two settings: when used as a standalone policy, where it outperforms competing learned policies, and as the policy within a MAPF solver with collision shielding, where it achieves the highest coverage on the MovingAI benchmark.
    \item We demonstrate the first learning-based MAPF policy to fully solve instances with over \textbf{one million} simultaneously acting agents in obstacle-rich environments.
\end{itemize}

\section{Related Work}

Existing MAPF approaches can be broadly divided into classical methods, which rely on predefined planning or coordination procedures, and learning-based methods, which acquire their decision rules from data. We first review classical approaches spanning explicit search, optimization, and reactive coordination, then turn to learning-based methods and their approaches to decentralized coordination.

\subsection{Classical MAPF Approaches}

Conflict-Based Search (CBS)~\citep{sharon2015conflict} and its improved variants~\citep{boyarski2015icbs,li2019disjoint} are canonical search-based MAPF methods. They systematically explore the joint state space and can provide optimal or bounded-suboptimal guarantees, but are often limited in scalability.

Reduction-based approaches reformulate MAPF as an equivalent well-studied optimization problem, such as minimum-cost flow or Boolean satisfiability (SAT), and use existing solvers to compute optimal or near-optimal solutions~\citep{surynek2016efficient,lam2022branch}.

Fast rule-based solvers such as PIBT~\citep{okumura2022priority} instead rely on simple local coordination rules to achieve high scalability, though they generally sacrifice optimality. Building on PIBT, LaCAM~\citep{okumura2023lacam,okumura2024engineering} integrates PIBT as a low-level policy within a search-based framework, combining reactive local decisions with conflict-aware global reasoning to improve solution quality. Similarly, MAPF-LNS2~\citep{li2022mapf} employs large neighborhood search with adaptive repair strategies for rapid generation and refinement of near-optimal solutions. Simpler approaches like prioritized planning~\citep{ma2019searching} trade optimality for runtime efficiency and remain popular in large-scale MAPF scenarios due to their computational simplicity.

These classical approaches make different trade-offs between solution quality, computational cost, and scalability. Search-based methods can provide stronger guarantees but typically incur increasing computational cost as the number of agents grows, while reactive and prioritized methods achieve greater scalability
through more local decision-making.

\subsection{Learning-Based MAPF Approaches}

Learning-based approaches learn coordination policies from data, offering an
alternative to the predefined planning and coordination procedures used by
classical MAPF methods. They differ in whether an agent commits to its action
in one shot. Methods with \textit{direct commitment} select an action from a
predicted distribution in one step, with any resulting conflicts handled
separately or left unresolved; methods with \textit{refinement before
commitment} revise an intermediate action choice over several passes before
acting.

\paragraph{Direct Commitment.}
One of the pioneering works, PRIMAL~\citep{sartoretti2019primal}, showed that decentralized agents using a learned policy could solve MAPF when the only shared information is the agents' targets, without any further inter-agent communication. Also without inter-agent communication, MAPF-GPT~\citep{andreychuk2025mapf} instead uses a Transformer-based architecture trained via imitation learning on a large dataset of expert MAPF solutions. MAPF-GPT-DDG~\citep{andreychuk2025advancing} further fine-tunes MAPF-GPT on additional expert data collected via active learning. SILLM~\citep{jiang2025deploying} scales imitation learning to lifelong MAPF with 10,000 agents. Concurrently with this work, PRIMAL3~\citep{he2026primal3} also targets scale, reporting single-instance stress tests with up to 100,000 agents at 20\% obstacle density.

To enable richer coordination, DHC~\citep{ma2021distributed} brought learned communication~\citep{sukhbaatar2016learning, foerster2016learning} to MAPF, with agents exchanging latent representations, improving performance over PRIMAL. DCC~\citep{ma2022learning} refines this with selective communication, deciding when and what to communicate to limit redundancy while preserving necessary information.

A related line of work uses graph attention to structure communication. MAGAT~\citep{li2021message} replaces a fixed communication structure with learned attention over neighboring agents, letting each agent weight incoming messages by relevance, and is trained via imitation learning on expert demonstrations, similar to MAPF-GPT. MAGAT+~\citep{jain2026graph} extends this with three stacked attention layers in place of the single layer used in the original, and adopts a two-stage imitation-learning paradigm: it is first pretrained on trajectories from a search-based expert, then fine-tuned with additional imitation data on the target map. HMAGAT~\citep{jain2026pairwise} instead replaces the pairwise graph with a hypergraph representation to capture group-level interactions among several neighboring agents at once, likewise trained purely by imitation learning.

A different family of methods builds communication into a Transformer rather than a graph attention mechanism. SCRIMP~\citep{wang2023scrimp} keeps a separate convolutional observation encoder and fuses neighboring agents' messages through a dedicated Transformer-based communication block. LC-MAPF~\citep{lcmapf2026} instead uses a Transformer encoder-decoder architecture for the whole pipeline, exchanging local latent representations over multiple communication rounds before each agent samples a single final action.

Across these approaches, communication enriches what each agent's policy conditions on, but the final action is sampled once, independently, from each agent's distribution after communication ends: even a multimodal policy can then recombine locally valid choices into an infeasible \mbox{joint action}.

Cooperative multi-agent reinforcement learning also recognizes that
independent per-agent policies cannot express coordinated joint behavior,
but existing remedies either relax decentralization or give up correlation at
execution: \citet{fu2022revisiting} order agents autoregressively and broadcast
each action to successors, AgentMixer~\citep{li2025agentmixer} correlates
policies only in training, and MACPF~\citep{wang2023more} recovers an
equally valued factorizable policy with a single mode. DMM does neither.

\paragraph{Refinement Before Commitment.}
Iterative refinement is the mechanism generative models such as diffusion use
to represent multimodal distributions, and a recent line brings it to
multi-agent planning. In discrete MAPF, DiffLNS~\citep{wang2026discrete} is a
concurrent example: it centrally refines the joint action tensor of all agents
into a full-horizon plan and passes the result to LNS2 for repair, motivated,
like this work, by the multimodality of the expert distribution. In contrast,
DMM performs decentralized, per-timestep intent refinement through
communication and commits the resulting actions directly.

A larger body of related work considers continuous-space multi-robot motion
planning, where refinement is likewise centralized in most cases
~\citep{liang2025simultaneous, shaoul2025multirobot, liang2026discrete}, although
decentralized variants exist. In these variants, coordination is introduced
separately from refinement: by inferring or simulating teammates while
refining alone~\citep{zhu2024madiff, liang2026simulation}, by a critic that
couples agents only during training~\citep{li2026diffusing}, or by exchanging
already planned trajectories~\citep{liang2026simulation, lew2026aid}. Thus,
communication typically provides either context for refinement or a decision
already formed. DMM instead communicates the intermediate refinement state
itself, allowing agents to condition on one another's intents while they are
still forming.

\section{Background}
\label{sec:problem}
 
\subsection{MAPF Preliminaries}

A MAPF instance is a tuple
$\bigl(G,\{s_u\}_{u\in U},\{g_u\}_{u\in U}\bigr)$, where
$G=(V,E)$ is a four-connected grid, $U=\{u_1,\ldots,u_n\}$ is the set of
agents, and $s_u,g_u\in V$ are the start and goal vertices of agent $u$,
respectively. Starts are pairwise distinct, as are goals. Time is discrete.
The (joint) configuration at timestep $t$ is
$\mathbf{v}^t=(v_u^t)_{u\in U}\in V^n$, with
$\mathbf{v}^0=(s_u)_{u\in U}$. At each timestep, each agent selects an action
from the common action set
$\mathcal{A}=\{\mathrm{wait},\mathrm{up},\mathrm{down},
\mathrm{left},\mathrm{right}\}$, forming the joint action
$\mathbf{a}^t=(a_u^t)_{u\in U}\in\mathcal{A}^n$.

A joint action is \emph{feasible} at $\mathbf{v}^t$ if every move is either a
wait action or traverses an edge in $E$, no two agents occupy the same vertex
after the transition, and no two agents traverse the same edge in opposite
directions during the same timestep. We denote the set of feasible joint actions at configuration
$\mathbf{v}$ by $\mathcal{F}(\mathbf{v})\subseteq\mathcal{A}^n$.

A solution is a sequence of feasible joint actions that reaches
$v_u^T=g_u$ for all $u\in U$ for some $T\leq H$, where $H$ is the execution
horizon.

We evaluate solution quality using the \emph{sum of costs} (SoC) and
\emph{makespan} (MS). Let $c_u$ denote the earliest timestep at which agent
$u$ reaches its goal and remains there for the remainder of the episode. The
SoC measures the total arrival cost across all agents, whereas the makespan
measures the arrival time of the last agent:
\[
\mathrm{SoC}=\sum_{u\in U}c_u,
\qquad
\mathrm{MS}=\max_{u\in U}c_u.
\]
We additionally report the \textit{success rate} (SR), defined as the fraction
of instances for which all agents reach their goals within $H$ timesteps.
 
\subsection{Decentralized MAPF with communication}

We model decentralized MAPF as a finite-horizon
Dec-POMDP~\citep{bernstein2002complexity}, defined by the tuple
\mbox{$M=\langle S,\mathcal{A},U,P,R,O,\mathcal{O}\rangle$},
where a state $s^t\in S$ comprises the agent configuration $\mathbf{v}^t$ and
the fixed targets. Given the current state and an executed joint action, the
transition function $P$ determines the next state. The reward function
$R:S\times\mathcal{A}^n\rightarrow\mathbb{R}$ assigns a single scalar reward
shared by all agents. Rewards are terminal and undiscounted.

At timestep $t$, agent $u$ receives the local observation
$o_u^t=\mathcal{O}_u(s^t)\in O$, consisting of an egocentric
$(2\rho+1)\times(2\rho+1)$ patch centered at $v_u^t$. The Dec-POMDP is
augmented with a communication channel represented by a dynamic directed graph
$G_{\mathrm{comm}}^t=(U,E_c^t)$. Each agent $u$ communicates with its $k$ nearest agents
within its observation range, including itself, denoted $\mathcal{N}^t(u)$. 

Each timestep consists of $K$ synchronous communication rounds followed by a
single action commitment. For \mbox{$r=1,\ldots,K$}, let
\begin{equation}
    x_u^{t,r}
    =
    \bigl(
        o_u^t,\,
        \{m_w^{t,r'}\}_{w\in\mathcal{N}^t(u),\,r'<r}
    \bigr)
    \label{eq:info}
\end{equation}
denote the information available to agent $u$ at the start of communication
round $r$, comprising its local observation $o_u^t$ and the messages
$m_w^{t,r'}$ received from its neighbors in preceding rounds. In particular,
$x_u^{t,1}=(o_u^t,\emptyset)$. Agent $u$ then emits
\begin{equation}
    m_u^{t,r}\sim\mu_\theta\bigl(\cdot\mid x_u^{t,r}\bigr),
    \label{eq:msg-def}
\end{equation}
where $\mu_\theta$ is the message-generation distribution shared by all agents;
it may be deterministic, as in conventional learned communication, or depend
on randomness private to agent $u$, as in DMM.
Messages are generated simultaneously in each round and subsequently
transmitted along $G_{\mathrm{comm}}^t$. After the $K$ rounds, let $x_u^t=x_u^{t,K+1}$
denote the information available to agent $u$ at action commitment. All agents
share the parameters of a stochastic policy
$\pi_\theta(a_u^t\mid x_u^t)$ and select their actions simultaneously.

Decentralized execution is required to satisfy:
\begin{enumerate}
    \item[(i)] Agent $u$ conditions its decision only on $x_u^t$.
    \item[(ii)] No agent observes another agent's committed action
    before selecting its own, and no agent ordering is imposed.
    \item[(iii)] No component encodes the global state $s^t$, predicts
the joint action $\mathbf{a}^t$, or directly aggregates global information;
communication is restricted to local neighbors.
\end{enumerate}

Under (i)--(iii), per-agent computation is independent of the number of agents
$n$, and a decentralized solution reduces to a collection of local policies;
communication augments $x_u^t$ but leaves the transition function $P$ and
reward $R$ unchanged.

\subsection{Training Objectives}

Let
\[
\mathcal{D}_E
=
\left\{
\left(
\{o_{u,j}\}_{u\in U_j},
\{a_{u,j}^{\star}\}_{u\in U_j},
G_{\mathrm{comm},j}
\right)
\right\}_{j=1}^{N_E}
\]
denote an expert demonstration dataset of $N_E$ timestep samples, where
$U_j$ is the set of agents in sample $j$, $o_{u,j}$ is agent $u$'s local
observation, $a_{u,j}^{\star}$ is its expert action, and
$G_{\mathrm{comm},j}$ specifies the communication graph. For a given sample,
the communication process defined above induces the context $x_{u,j}$
available to agent $u$ at action commitment.

For conventional action-based imitation, the shared policy is trained by
minimizing the per-agent cross-entropy
\begin{equation}
    \mathcal{L}_{\mathrm{CE}}(\theta)
    =
    -\frac{1}{N_E}
    \sum_{j=1}^{N_E}
    \frac{1}{|U_j|}
    \sum_{u\in U_j}
    \log
    \pi_\theta
    \left(
        a_{u,j}^{\star}\mid x_{u,j}
    \right).
    \label{eq:cross-entropy}
\end{equation}
This objective matches each agent's action distribution to the corresponding
expert action conditioned on the information available at commitment.

Under independent final-action sampling, each agent samples its action from
the shared policy conditioned on its local context after the $K$
communication rounds. For a fixed timestep, suppressing the sample and time
indices for clarity, the resulting joint-action distribution factorizes as
\begin{equation}
Q_{\pi_\theta}(\mathbf{a}\mid x)
=
\prod_{u\in U}\pi_\theta(a_u\mid x_u),
\qquad
x=(x_1,\ldots,x_n).
\label{eq:product-factorization}
\end{equation}
Thus, communication may affect each agent's action distribution through its
local context $x_u$, but under independent final-action sampling the action
choices remain conditionally independent given $x$.

A reinforcement-learning objective instead optimizes the expected team return
over decentralized rollouts. Let
\[
\zeta=(s^0,\mathbf{a}^0,s^1,\mathbf{a}^1,\ldots,s^T)
\]
denote a rollout, where $T\leq H$ is the termination timestep, reached when
the joint goal configuration is achieved or the horizon is exhausted. Given
the terminal team return $R(\zeta)\in\mathbb{R}$, the objective is
\begin{equation}
    J(\theta)
    =
    \mathbb{E}_{\zeta\sim\pi_\theta}
    \left[
        R(\zeta)
    \right].
    \label{eq:joint-return}
\end{equation}

\section{Method}
\label{sec:method}

\subsection{The Decentralized Factorization Gap}
\label{sec:gap}

Communication can give each agent a richer local context, but conventional
decentralized policies still sample their final actions independently. This
last step can discard information about which individual choices belong
together. Consequently, even if every agent learns its expert action
distribution exactly, the sampled joint action need not follow the expert
joint distribution.

Let $P^\star$ denote the expert distribution. Let
$A=(A_1,\ldots,A_n)$ be the expert joint action and
$X=(X_1,\ldots,X_n)$ the information available at action commitment, where
$X_u$ is the local information available to agent $u$ and $X_{-u}$ denotes the
remaining context. For an independently sampling decentralized executor, let
$q_u(a_u\mid x_u)$ denote the action distribution used by agent $u$ given its
local information. Its joint distribution has the product form
\[
Q_q(a\mid x)=\prod_{u\in U}q_u(a_u\mid x_u).
\]
An analogous restriction appears in non-autoregressive sequence models, which
predict output tokens independently in parallel. There, independently learned
token marginals can mix parts of different valid translations, which is known as the
``multimodality problem''~\citep{gu2018nonautoregressive}. \citet{huang2022learning} formalized the resulting information loss as a
KL lower bound given by conditional total correlation. We transfer this
argument from token positions to agents; because each agent observes only its
own $X_u$, the decentralized case has an additional penalty for missing local
information.

The dependence among expert actions that remains after the full context is
known is measured by the conditional total
correlation~\citep{watanabe1960information}
\[
\mathrm{TC}(A\mid X)
:=
\mathbb{E}_{X}D_{\mathrm{KL}}\!\left(
P^\star(A\mid X)
\,\middle\|\,
\prod_{u\in U}P^\star(A_u\mid X)
\right).
\]
For a fixed context $x$, the KL term is zero exactly when the expert joint
distribution factorizes into its per-agent marginals. Hence,
$\mathrm{TC}(A\mid X)>0$ whenever the expert actions remain dependent after
fixing $X$ with nonzero probability. Intuitively, knowing one agent's expert
action then provides information about the others. This occurs, for example,
when the expert assigns positive probability to several coordinated joint
actions but not to all recombinations of their individual actions. In the
corridor, with $g$ for ``go'' and $w$ for ``wait,'' $(g,w)$ and
$(w,g)$ are selected, but
$(g,g)$ and $(w,w)$ are not.

\paragraph{Proposition (decentralized factorization gap).}
Suppose $\mathrm{TC}(A\mid X)>0$. Even allowing each local policy arbitrary
capacity, the best product executor satisfies
\begin{equation}
\begin{aligned}
\mathcal{G}_{\mathrm{dec}}
&:=\min_{\{q_u\}_{u\in U}}
\mathbb{E}_{X}D_{\mathrm{KL}}\!\left(
P^\star(A\mid X)\,\middle\|\,Q_q(A\mid X)
\right) \\
&=\mathrm{TC}(A\mid X)
+\sum_{u\in U}I(A_u;X_{-u}\mid X_u) \\
&\geq \mathrm{TC}(A\mid X)>0.
\end{aligned}
\label{eq:decentralized-factorization-gap}
\end{equation}
The minimum is attained by
$q_u^\star(A_u\mid X_u)=P^\star(A_u\mid X_u)$, the Bayes-optimal solution of
per-agent cross-entropy training. The proof is given in \appref{sec:factorization-gap}.
Thus, Eq.~\eqref{eq:decentralized-factorization-gap}
describes an irreducible error rather than imperfect learning. The first term
is the cost of discarding residual dependence between agents' actions. The
second is the cost of action-relevant information that exists in $X_{-u}$ but
is unavailable in $X_u$. Positive total correlation is sufficient, but not
necessary, for a strict gap: missing local information can also make the gap
positive when $\mathrm{TC}(A\mid X)=0$.
If every factor instead observes the same full context, $X_u=X$ for all $u$,
the mutual-information terms vanish and
Eq.~\eqref{eq:decentralized-factorization-gap} reduces to the
non-autoregressive Transformer bound of \citet{huang2022learning}.
Related separations between independent and correlated policies have been
shown for value decomposition~\citep{fu2022revisiting} and for attainable
return~\citep{wang2023more}; Eq.~\eqref{eq:decentralized-factorization-gap}
instead concerns matching the expert joint distribution, the target of
per-agent imitation, and additionally accounts for information missing from
each agent's local context.

The corridor example makes the first term concrete. At its ambiguous context
$x$, suppose the expert chooses the
two coordinated resolutions with equal probability,
\[
P^\star((g,w)\mid x)=P^\star((w,g)\mid x)=\tfrac12.
\]
Each exact local marginal is then uniform over $\{g,w\}$. Independent sampling
therefore assigns probability $1/4$ to every pair, including both the
collision $(g,g)$ and the stall $(w,w)$. At this context,
\[
\mathrm{TC}(A\mid X=x)
=2\left(\frac12\log\frac{1/2}{1/4}\right)
=\log 2>0.
\]
Nothing is wrong with either local marginal; the error appears only when they
are multiplied. Deterministic communication can reduce the missing-information term by moving
more of $X_{-u}$ into each $X_u$. Once the communication transcript is fixed,
however, independent final sampling still produces a product distribution and
cannot represent residual action dependence. Escaping the product form
requires that the exchanged information itself carry the agents' stochastic
choices, so that each agent can condition on its neighbors' samples before
committing. We therefore seek a decentralized
mechanism that preserves local, simultaneous execution while allowing agents
to coordinate their sampled choices before commitment.

\subsection{DMM: Decentralized Iterative Intent Refinement}

DMM augments a decentralized communication policy with an \emph{action intent}
that is refined during communication and included in the messages exchanged
with neighboring agents. Each agent encodes its local observation into a
compact representation $\ell_u$, which remains fixed across refinement rounds,
while the communicated intent evolves across rounds. The experiments
instantiate this mechanism with two network architectures, DMM-3M and
DMM-0.8M, described in \appref{sec:dmm-architectures}.

The intent update could be trained as a standard refinement step, as in
diffusion or flow matching. However, DMM couples refinement with learned
communication: each round's state determines the message passed to the next
round. This prevents independent training of refinement steps and motivates
the sequential communication-refinement process described next.

\paragraph{Intent initialization.}
Agent $u$ maintains an intent
$z_u^r\in\mathbb{R}^{|\mathcal{A}|}$, with one value for each action. Rather
than representing a probability vector directly, the intent is maintained in
mean-centered log-space, which preserves relative action preferences while
removing their arbitrary common offset. At inference, the initial intent is
sampled independently from an uninformative Dirichlet prior:
\begin{equation}
z_u^0=\mathrm{lc}(\log d_u),
\qquad
d_u\sim\mathrm{Dirichlet}(\mathbf{1}),
\label{eq:z0}
\end{equation}
where $\mathrm{lc}(v)=v-\bar v$ denotes mean-centering.

\begin{figure}[tb!]
    \centering
    \includegraphics[width=1.0\linewidth]{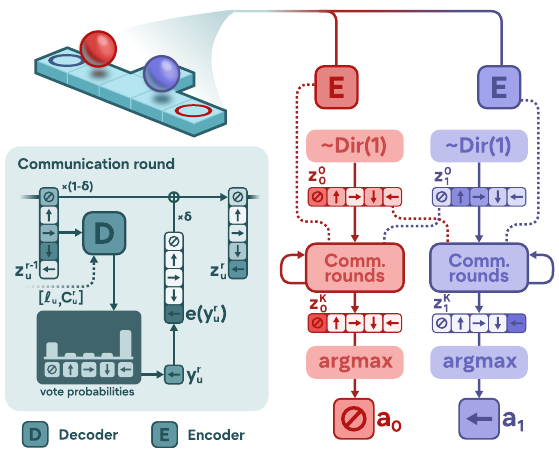}
    \caption{%
    Overview of DMM for two agents resolving a shared conflict (Section~\ref{sec:method}). Each agent encodes
    its observation into a latent and initializes an action intent, then exchanges
    intent-augmented messages over $K$ rounds while refining its intent toward
    successive votes. In this example, the resulting actions are jointly consistent: the red agent waits while the blue agent moves left.
    }
    \label{fig:dmm_architecture}
\end{figure}

\paragraph{Iterative refinement.}
Figure~\ref{fig:dmm_architecture} illustrates the iterative communication and
refinement process. At each round $r\in\{1,\ldots,K\}$, agent $u$ broadcasts its
current intent together with a learned message feature:
\begin{equation}
m_u^r=h_u^{r-1}+W_s z_u^{r-1},
\label{eq:message}
\end{equation}
where $h_u^{r-1}$ is the learned message feature and $W_s$ projects the intent
into the message space. Agent $u$ then receives messages from its local
neighbors and processes them together with its fixed latent representation
$\ell_u$. The decoder uses this local neighborhood context to produce action
logits $\phi_u^r$ and the next message feature $h_u^r$:
\begin{equation*}
\begin{gathered}
    C_u^r=
    \bigl\{
        m_v^r
    \bigr\}_{v\in\mathcal{N}(u)\cup\{u\}}, \\
    \psi_u^r=
    \mathrm{Decoder}(\ell_u,C_u^r),\\
    \phi_u^r=\mathrm{PolicyHead}(\psi_u^r),
    \qquad
    h_u^r=\mathrm{MsgHead}(\psi_u^r).
\end{gathered}
\end{equation*}
From the action logits, the agent samples a discrete vote:
\begin{equation}
p_{\theta,u}^r
=
\mathrm{softmax}(\phi_u^r), \quad
y_u^r\sim
\mathrm{Categorical}
\left(
p_{\theta,u}^r
\right).
\end{equation}
The sampled vote is then incorporated into the current intent:
\begin{equation}
z_u^r
=
(1-\delta)z_u^{r-1}
+
\delta e(y_u^r),
\label{eq:z_update}
\end{equation}
where
$e(y)=\mathrm{lc}(\log(\mathrm{onehot}(y)+\eta))$
with $\eta>0$ a small constant that prevents taking the logarithm of zero. The
embedding places the vote in the same centered log-space as the intent. The step size
$\delta\in(0,1)$ retains part of the previous intent while incorporating the
current vote, so the intent accumulates information across refinement rounds.
Because the vote is discrete, gradients do not pass through the vote or the
intent update. The message feature remains differentiable, however, allowing
later-round losses to influence earlier communication.

The key idea in DMM is that the quantity being refined is also part of the
communication signal. Each agent therefore observes its neighbors' evolving
intents and can adapt its own subsequent votes before commitment. Although the
initial intents are sampled independently, the refinement trajectories become
coupled through communication, allowing different runs to settle on different
coordinated joint-action outcomes.
After the $K$ refinement rounds, each agent commits to the action favored by
its own final intent:
\begin{equation}
a_u^t=\arg\max_{a\in\mathcal{A}}z_u^K[a].
\end{equation}
Agents therefore coordinate through their exchanged refinement states before
commitment, while each agent selects its final action locally from its own
refined intent.

\subsection{Imitation Pretraining}

DMM is first pretrained on expert trajectories by supervising the votes
at every refinement round.
For an expert action $a_u^{t,\star}$, the imitation loss is
\begin{equation*}
    \mathcal{L}_{\mathrm{IL}}
    =
    -\frac{1}{K}
    \sum_{r=1}^{K}
    \log
    p_{\theta,u}^r
    \left(
    a_u^{t,\star}
    \right)
\end{equation*}

Early in training, however, noisy model votes produce uninformative intents
that are then communicated to neighboring agents. We therefore use
two forms of teacher forcing. With probability $\beta_0$, the initial intent
is biased toward the expert action by sampling
\begin{equation}
    d_u\sim
    \mathrm{Dirichlet}
    \left(
        \mathbf{1}
        +\mathrm{onehot}(a_u^{t,\star})
    \right),
\end{equation}
instead of the uninformative prior in Eq.~\eqref{eq:z0}. With probability
$\beta_r$ at each round, the sampled vote $y_u^r$ is replaced by the expert
action before applying Eq.~\eqref{eq:z_update}. The first intervention provides
an informative initial condition; the second keeps the evolving intent
trajectory informative for neighboring agents.
The latter is particularly relevant to DMM because the resulting intent is
communicated in subsequent rounds, so errors in early votes can also alter the
context received by neighboring agents.
Both probabilities are annealed during pretraining toward a retained nonzero
floor, reducing the amount of teacher forcing while preserving an expert
signal in the evolving intent trajectory. The teacher-forcing schedules used
in our experiments are specified in Section~\ref{sec:experimental-setup}.

Imitation teaches local agreement with coordinated expert actions but does not directly
optimize the quality of the resulting joint solution. We therefore fine-tune the
same refinement process directly on task-level outcomes.

\subsection{MICPO: Critic-Free Multi-Agent Fine-Tuning}

We use reinforcement learning to optimize the complete decentralized rollout
with a shared team objective, reflecting MAPF's cooperative nature, since each
agent's actions can affect the outcomes of others. Conventional actor-critic
methods are poorly matched to this setting: a centralized critic must
generalize over a combinatorially large joint state, while a decentralized
critic has only partial information about the global outcome. MICPO therefore
removes the value function and estimates advantage by comparing rollouts under
matched conditions, adapting group-relative optimization~\citep{deepseek-math}
to DMM's multi-agent, multi-round setting.
Group-relative optimization has also been applied to multi-agent LLM
collaboration~\citep{liu2026llm} and refined with step-level groups of
recurring anchor states~\citep{feng2025gigpo}. MICPO differs in constructing
groups that share the sampled initial intent at every step and in computing
importance ratios per agent and per refinement round.

\paragraph{Matched rollout groups.}
At each optimization iteration, we sample $B$ scenarios and construct $M$
matched groups of $G$ trajectories per scenario. At every environment step
$t$, the $G$ trajectories within a group share the same sampled initial intent
$z_t^0$, while the subsequent refinement votes are sampled independently.
Thus, trajectories in a group are matched on the scenario and on $z_t^0$ at
each step, while their states may diverge due to earlier
stochastic refinement decisions.

\paragraph{Team return and group-relative advantage.}
For rollout $\zeta$, let $H_\zeta$ denote its number of environment
transitions, and define the off-goal duration of agent $u$ as
\begin{equation}
T_{\mathrm{off},u}(\zeta)
=
\sum_{t=0}^{H_\zeta-1}
\mathbf{1}\!\left[v_u^t \neq g_u\right].
\end{equation}
Let $B_u(\zeta)$ denote the number of blocked non-wait actions proposed by
agent $u$ during the rollout. We form two trajectory-level components:
\begin{align}
q_c(\zeta)
&=
-\frac{1}{|U|}
\sum_{u\in U}
\log\!\left(1+T_{\mathrm{off},u}(\zeta)\right),
\\
q_b(\zeta)
&=
-\frac{1}{|U|}
\sum_{u\in U}
B_u(\zeta).
\end{align}
Their weighted sum defines the unnormalized team return
\begin{equation}
R(\zeta)
=
w_c q_c(\zeta)
+
w_b q_b(\zeta).
\label{eq:micpo-return}
\end{equation}

Because the two components have different natural scales, we normalize them
separately within each matched group of $G$ trajectories, as in
GDPO~\citep{liu2026gdpogrouprewarddecouplednormalization}. We define
\begin{equation}
N_G(x)
=
\begin{cases}
\dfrac{x-\mu_G(x)}{\sigma_G(x)},
    & \sigma_G(x)\geq\tau, \\[6pt]
0,
    & \sigma_G(x)<\tau,
\end{cases}
\end{equation}
where $\mu_G(x)$ and $\sigma_G(x)$ are the mean and standard deviation within
the matched group, and $\tau>0$ is a small numerical-stability threshold.
When the group standard deviation falls below this threshold, the normalized
value is set to zero to avoid division by a near-zero quantity. The
group-relative advantage is
\begin{equation}
\hat A(\zeta)
=
N_G\!\left(
    w_c N_G\!\left(q_c(\zeta)\right)
    +
    w_b N_G\!\left(q_b(\zeta)\right)
\right).
\label{eq:micpo-advantage}
\end{equation}
The resulting advantage is shared across all agents and communication rounds
of the corresponding rollout.

\paragraph{Bounded replay.}
Replaying every collected decision is expensive because MAPF episodes can be
long and each environment step contains $K$ refinement rounds. We therefore
rank trajectories separately within each matched group according to the team
return $R(\zeta)$ and retain the $\kappa$ highest- and $\kappa$
lowest-return trajectories. The group-relative advantages in
Eq.~\eqref{eq:micpo-advantage} are computed from all $G$ trajectories before
this filtering, so replay selection does not change the comparison group used
for credit assignment. From each retained trajectory, we sample $S$
environment timesteps for optimization. Full replay and sampling details are
given in \appref{sec:micpo-details}.

\paragraph{Round-level policy optimization.}
Because each DMM action is produced from $K$ stochastic refinement votes,
MICPO applies policy optimization at the level of individual votes. Recall that
$p_{\theta,u}^r=\mathrm{softmax}(\phi_u^r)$
denotes the round-$r$ action distribution of agent $u$. We similarly define
\[
p_{\mathrm{old},u}^r
=
\mathrm{softmax}(\phi_{u,\mathrm{old}}^r),
\qquad
p_{\mathrm{ref},u}^r
=
\mathrm{softmax}(\phi_{u,\mathrm{ref}}^r)
\]
for the old and reference policies, respectively. MICPO computes an importance
ratio separately for each agent and refinement round rather than collapsing
the refinement process into a single episode-level ratio:
\begin{equation}
\rho_u^r
=
\frac{
    p_{\theta,u}^r(y_u^r)
}{
    p_{\mathrm{old},u}^r(y_u^r)
}.
\end{equation}
For each selected environment timestep, all $K$ refinement rounds are replayed
together. The stored initial intent and sampled vote sequence reconstruct the
intent trajectory, while message features are recomputed across rounds so that
gradients propagate through the complete communication pathway.

The clipped objective is averaged over agents and rounds:
\begin{equation*}
\mathcal{L}_{\mathrm{clip}}
=
-\mathbb{E}_{u,r}
\left[
    \min
    \left(
        \rho_u^r \hat A,\,
        \mathrm{clip}
        \left(
            \rho_u^r,
            1-\varepsilon,
            1+\varepsilon
        \right)
        \hat A
    \right)
\right].
\end{equation*}

To limit drift from the coordinated behavior learned during imitation
pretraining, we additionally regularize the policy toward a frozen reference
policy initialized from the imitation-pretrained checkpoint, similar
to~\citep{deepseek-math}. The reference-policy penalty is evaluated at each
agent and communication round:
\begin{equation*}
\mathcal{L}_{\mathrm{KL}}
=
\alpha_{\mathrm{KL}}
\mathbb{E}_{u}
\left[
    \sum_{r=1}^{K}
    D_{\mathrm{KL}}
        \left(
            p_{\theta,u}^r
            \,\middle\|\,
            p_{\mathrm{ref},u}^r
        \right)
\right].
\end{equation*}
The final MICPO objective is $\mathcal{L}=\mathcal{L}_{\mathrm{clip}}+\mathcal{L}_{\mathrm{KL}}$.

The reference policy anchors each round's action distribution to the
imitation-pretrained behavior, while the communication pathway remains
trainable, allowing fine-tuning to reshape communication while limiting drift
from the coordination learned during imitation.

\section{Experimental Setup}
\label{sec:experimental-setup}

Our evaluation combines two benchmark studies with two targeted experiments.
POGEMA~\citep{skrynnik2025pogema} evaluates learned policies directly under
controlled increases in team size, while MovingAI~\citep{stern2019multi}
evaluates MAPF methods with their associated action-processing mechanisms
across diverse maps and agent counts. We
additionally evaluate large-scale execution up to one million agents and use a
corridor scenario to examine the joint-action distribution induced
by refinement.

\paragraph{DMM configurations and training.}
We evaluate two implementations of the same DMM intent-refinement mechanism.
DMM-3M retains the LC-MAPF Transformer encoder--decoder backbone~\citep{lcmapf2026}, enabling a
controlled comparison with LC-MAPF-3M. DMM-0.8M is a compact implementation
that moves most observation-dependent computation outside the refinement loop
and reuses it across rounds, reducing the computation repeated during
refinement. It contains 763,296 trainable parameters, approximately
$4.25\times$ fewer than DMM-3M, and is used as the computationally lighter
configuration in the large-scale experiments. Both implementations preserve
the same action-intent representation, local communication semantics, and
final action rule; full architectural details are given in
\appref{sec:dmm-architectures}.

Each agent receives a tokenized observation of up to 256 tokens that encodes
its \(11\times11\) local field of view, agent-specific attributes, and spatial
context. Agents can communicate with up to 13 nearby agents within a 5-cell
radius, including themselves (i.e., up to 12 neighbors).

The imitation-pretraining dataset was derived from LC-MAPF~\citep{lcmapf2026}
and consists of aggregated samples from \textit{mazes}, \textit{random}, and
\textit{house} maps with up to 32 agents (a 0.6:0.2:0.2 split); the house map
generator follows~\citep{he2024alpha}. Each sample includes tokenized agent
observations, the corresponding ground-truth actions, and adjacency
information describing the local communication structure.

Both DMM variants are pretrained by imitation for 1,000,000 iterations and
subsequently fine-tuned with MICPO. During imitation pretraining, both
teacher-forcing probabilities are initialized at $1.0$ and annealed over the
first 100,000 iterations to the retained floor
$\beta_0=\beta_r=0.8$. DMM-3M uses AdamW with cosine learning-rate decay and
requires approximately 272.8 GPU-hours across four H100 GPUs. Its MICPO
fine-tuning runs for 500 outer iterations (96,000 optimizer updates) with
group size $G=24$, requiring 56.3 GPU-hours across four H100 GPUs.
DMM-0.8M is likewise pretrained for 1,000,000 iterations, with an effective
batch size of 800, requiring approximately 100.3 GPU-hours across four H100
GPUs. Its MICPO fine-tuning also runs for 500 outer iterations
(96,000 optimizer updates) with $G=24$, requiring approximately 14.1 GPU-hours
across four H100 GPUs. The complete architecture and training configurations
for both model sizes are provided in \appref{sec:hyper}.

\paragraph{POGEMA benchmark.}
We follow the LC-MAPF evaluation protocol~\citep{lcmapf2026} on the POGEMA
benchmark~\citep{skrynnik2025pogema}. We compare DMM against
HMAGAT~\citep{jain2026pairwise},
MAGAT+~\citep{jain2026graph}, MAPF-GPT-85M~\citep{andreychuk2025mapf},
MAPF-GPT-DDG-2M~\citep{andreychuk2025advancing}, and
LC-MAPF-3M~\citep{lcmapf2026} on \textit{Random}, \textit{Mazes},
\textit{Warehouse}, and \textit{Cities-Tiles} maps. Only learnable policies
are included in this comparison. Every method acts as a standalone policy,
without collision shielding, search, or other action-repair mechanisms, so the
results reflect the learned policies themselves rather than downstream
correction by an auxiliary planner.

Random and Mazes ($17\times17$ to $21\times21$, up to 96 and 80 agents,
respectively) use map families seen during pretraining, while Warehouse ($33\times46$,
up to 192 agents) and Cities-Tiles ($64\times64$, up to 256 agents) differ
in topology and agent count from the training set and are used to evaluate
out-of-distribution generalization. Episode length is capped at 128 steps,
except on Cities-Tiles, where it is 256. 

\paragraph{MovingAI benchmark.}
MovingAI~\citep{stern2019multi} contains 33 maps, each with 25 even and 25
random scenarios. For each scenario, we evaluate the instance with the maximum
available number of agents, ranging from 32 to 8,000 depending on the map. We
exclude only \texttt{maze-128-128-1}: preliminary experiments showed that none
of the evaluated methods could solve its maximum-agent instances. The reported
benchmark therefore contains 32 maps and 1,600 tasks.

We compare against the search-based LG-LaCAM~\citep{arita2026local} solver, which builds on the LaCAM
search structure~\citep{okumura2023lacam,okumura2024engineering};
MAPF-LNS2~\citep{li2022mapf}, which repeatedly replans subsets of conflicting
paths until it obtains a feasible solution; the hybrid LaGAT
solver~\citep{jain2026graph}, which combines the learned MAGAT+ policy with
LaCAM search; and the learned HMAGAT policy~\citep{jain2026pairwise}.
LG-LaCAM and LaGAT can explore alternative configurations through their search
structure, while MAPF-LNS2 can revise earlier planning choices through
neighborhood repair. Each search-based or hybrid method receives a wall-clock
budget of 600 seconds per task. HMAGAT and our DMM-MICPO-0.8M and DMM-MICPO-3M policies act directly in the
environment and receive a budget of 5,000 environment steps.

None of the evaluated methods performs post-solution refinement. Each run stops as soon as it finds a valid solution or exhausts its computation budget, in which case the task is counted as unsolved. A returned solution could later be optimized using LNS, but this stage is outside the scope of this work. We therefore evaluate the success rate, because an unsolved task does not provide a solution to optimize; time to a valid solution, because faster solving leaves more of a fixed overall budget for subsequent optimization; and solution quality, because the quality of the solution supplied to LNS can affect the result of time-limited refinement~\citep{arita2026local}.

All three learned policies use CS--PIBT, an established collision-shielding technique previously applied to learned MAPF policies~\citep{veerapaneni2025work,jain2026pairwise}. We additionally introduce Repeat-State Escape (RSE) for reactive rollouts. When CS--PIBT proposes a joint configuration that has already been visited, RSE temporarily forbids a responsible action and asks the policy and shield to construct a different successor before any action is executed. Both components use the global configuration and are therefore centralized, but RSE does not branch from past states or roll back executed actions. To assess RSE across decoding choices, we evaluate three configurations for each policy: sampling, sampling with RSE, and argmax with RSE. The main MovingAI comparison uses sampling with RSE for HMAGAT and argmax with RSE for both DMM variants. The full comparison and implementation details are given in \appref{sec:movingai-rse-comparison} and \appref{sec:movingai-extended}.

\begin{figure*}[t!]
    \centering
    \includegraphics[width=1.0\linewidth]{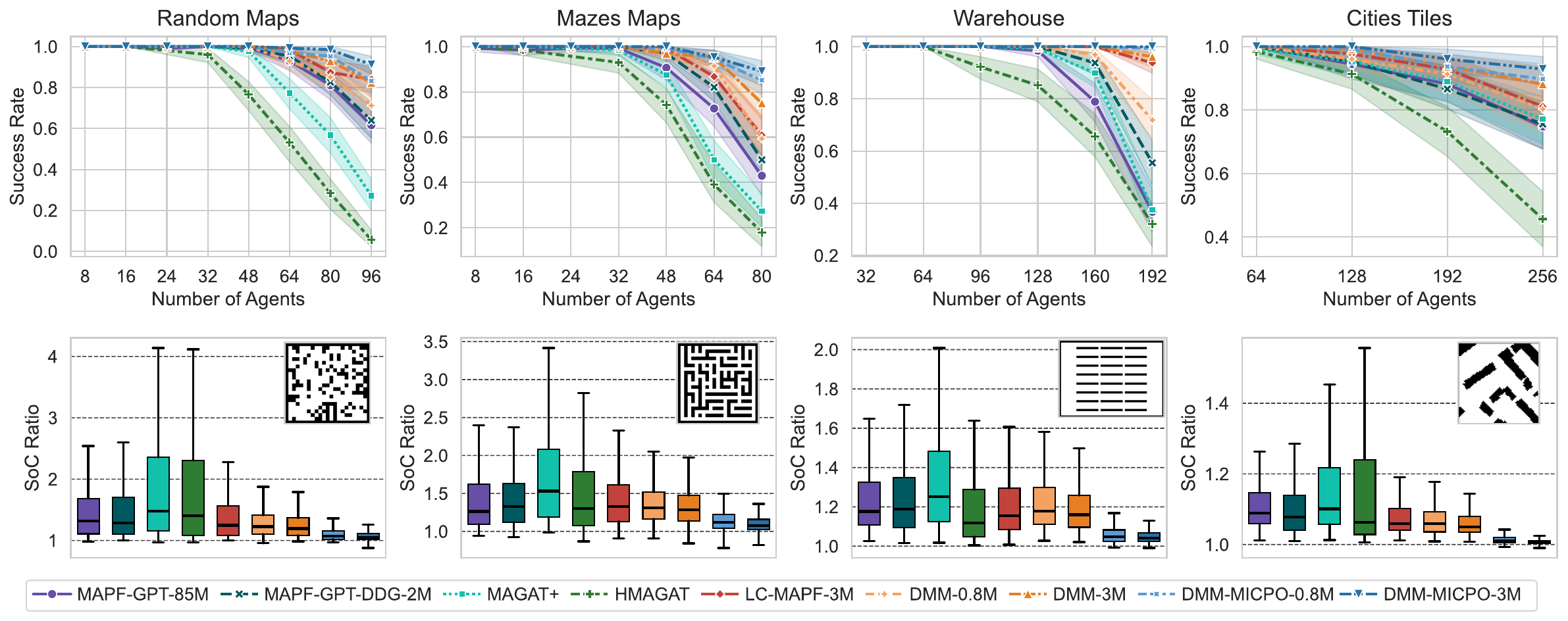}
    \caption{POGEMA benchmark results for four DMM variants (DMM-0.8M, DMM-3M,
        DMM-MICPO-0.8M, DMM-MICPO-3M) and five learnable baselines (MAPF-GPT-85M,
        MAPF-GPT-DDG-2M, MAGAT+, HMAGAT, LC-MAPF-3M), across four POGEMA domains:
        Random, Mazes, Warehouse, and Cities-Tiles.
        Top: success rate versus number of agents; each point is the mean over
        $n=128$ task instances per agent count, with shaded bands showing 95\%
        Wilson score CIs.
        Bottom: SoC ratio relative to the
        centralized LaCAM* solution; each box pools solved instances across
        \emph{all} evaluated agent counts for a given map. }
    \label{fig:success_rates}
\end{figure*}

\paragraph{Large-scale scalability experiment.}
We evaluate DMM-MICPO-0.8M on $2304\times2304$ maze maps with populations
from 131,072 to 1,048,576 agents. CS--PIBT shielding ensures
collision-free execution.

For DMM, we skip inference for an individual agent when it and all agents
in its local field of view are at their goals, proposing a wait instead.
We reevaluate this condition every step. CS--PIBT processes all proposed
actions, including skipped agents' waits, ensuring collision-free execution.

We run environment transitions and observation construction on the GPU,
including batched BFS with cached distances for cost-to-go observations
(see \appref{sec:gpu-pipeline}).
Our GPU implementation of CS--PIBT groups agents whose candidate moves may
conflict and processes independent groups in parallel. Within each group,
it follows standard PIBT priority ordering and backtracking, producing the
same joint action as standard sequential PIBT. However, at high density,
agents can still form one large conflict group, limiting parallelism and
making shielding a major runtime cost.

The same GPU-resident infrastructure supports GPU-PIBT, which uses
BFS-based move preferences instead of a learned policy. These preferences
are computed across GPUs and collected on one GPU for PIBT conflict
resolution, after which the selected actions are broadcast.

At each population size, we evaluate both methods on four different maze
layouts, using matched scenarios on $2304\times2304$ maps with approximately
29.1\% obstacles and an $H=32768$ horizon. Agent density ranges from
3.48\% to 27.86\% of traversable cells. Starts and goals are sampled
uniformly from traversable cells, with all start and goal cells distinct.
This differs from the prior million-agent evaluation
of~\citet{andreychuk2025advancing}, which used an empty $2048\times2048$ grid
with start--goal distances capped at 64.
Episodes end when all agents reach their goals or the horizon is reached.
We report means across the four scenarios, along with minimum and maximum
episode lengths. Mean step and PIBT times exclude the one-time BFS prefill;
PIBT time includes action broadcast. Total runtime includes prefill and
all subsequent computation and communication. Amortized decision time
divides total runtime by the number of agent-steps, including agents
whose policy inference was skipped.

\paragraph{Corridor experiment.}
To isolate the independent-sampling failure in which individually plausible
actions can be recombined into an incompatible joint action, we use a minimal
corridor scenario: a single-width passage with one side cell where an agent
can step aside. Two agents start at opposite ends of the corridor and must
swap positions to reach their goals (Figure~\ref{fig:corridor}). Two expert
trajectories solve the scenario, one in which the first agent yields at the
side cell and one in which the second agent does. The two trajectories share
exactly one state at which both resolutions remain possible. At this shared
state, two of the four combinations correspond to coordinated resolutions;
of the remaining two, mutual waiting produces a stall and simultaneous
movement produces a collision.

We train DMM, LC-MAPF, MAGAT+, and HMAGAT by imitation learning on the same
two expert trajectories, with each method using its native observation
representation. DMM is trained with $K=4$ refinement rounds and LC-MAPF with
four communication rounds, while MAGAT+ and HMAGAT retain the architectural
configurations of their original methods. All models in the corridor
experiment are trained for 5,000 iterations. For DMM, both teacher-forcing
probabilities are initialized at $1.0$ and annealed over the first 1,000
iterations to the same retained floor used in the main DMM training,
$\beta_0=\beta_r=0.8$. To examine the role of this retained signal, we
additionally train a zero-floor DMM variant in which both probabilities are
annealed from $1.0$ to $0$ over the same 1,000 iterations, with the
architecture and all other training settings unchanged. For each of five
training seeds, we sample 1,000 joint actions at the shared ambiguous state
from the trained policy.

\paragraph{Evaluation hardware.}
All evaluations were conducted on a server with two Intel Xeon Platinum
8480+ CPUs (56 cores each, 2.0--3.8 GHz), 2 TB of system
memory, and four NVIDIA H100 80 GB GPUs.
In all GPU-based evaluations other than the large-scale scalability
experiment, each worker used one GPU and processed independent instances. In the
large-scale experiment, each rollout was distributed across all four GPUs.

\section{Experimental Results}

\subsection{POGEMA Benchmark}
\label{sec:pogema-results}

Figure~\ref{fig:success_rates} reports success rate as a function of the number
of agents (top) and the distribution of sum-of-costs (SoC) ratios relative to
LaCAM* over solved instances pooled across all evaluated agent counts within
each domain (bottom), for DMM-0.8M, DMM-3M,
DMM-MICPO-0.8M, DMM-MICPO-3M, and five learnable baselines across the four
POGEMA domains.

As the number of agents grows, the success rates of all methods generally
decline, with the degradation becoming more pronounced at larger team sizes.
The gap between DMM-MICPO-3M and LC-MAPF-3M is largest on Warehouse and
Cities-Tiles: at the maximum
evaluated team size on each map (192 and 256 agents), DMM-MICPO-3M's success
rate remains at 1.000 on Warehouse and 0.922 on Cities-Tiles, while LC-MAPF-3M, the highest-success baseline at the maximum evaluated team
size in both domains, reaches 0.938 and 0.805, respectively.

The results highlight three properties of the proposed approach.
First, the comparison with LC-MAPF provides a controlled test of iterative
intent refinement. DMM-3M and LC-MAPF-3M share the same encoder-decoder
architecture, communication bottleneck, and training data at matched
parameter count, but differ in how the final action is produced: DMM uses
$K$ rounds of discrete intent refinement, whereas LC-MAPF uses direct
commitment sampling. DMM-3M generally achieves higher success rates than LC-MAPF-3M across agent
counts and lower pooled SoC ratios across domains, although the magnitude of
the difference varies across settings. In some settings the difference is
pronounced, particularly at larger team sizes, while in others the two
methods perform similarly with a modest advantage for DMM. This pattern is
consistent with a benefit from iterative refinement over direct commitment
sampling. Ablations over the number of inference-time refinement
rounds and over the content of the communicated messages are reported in
\appref{sec:round-ablation} and \appref{sec:z-ablation}.

\begin{figure}[!t]
    \centering
    \includegraphics[width=\linewidth]{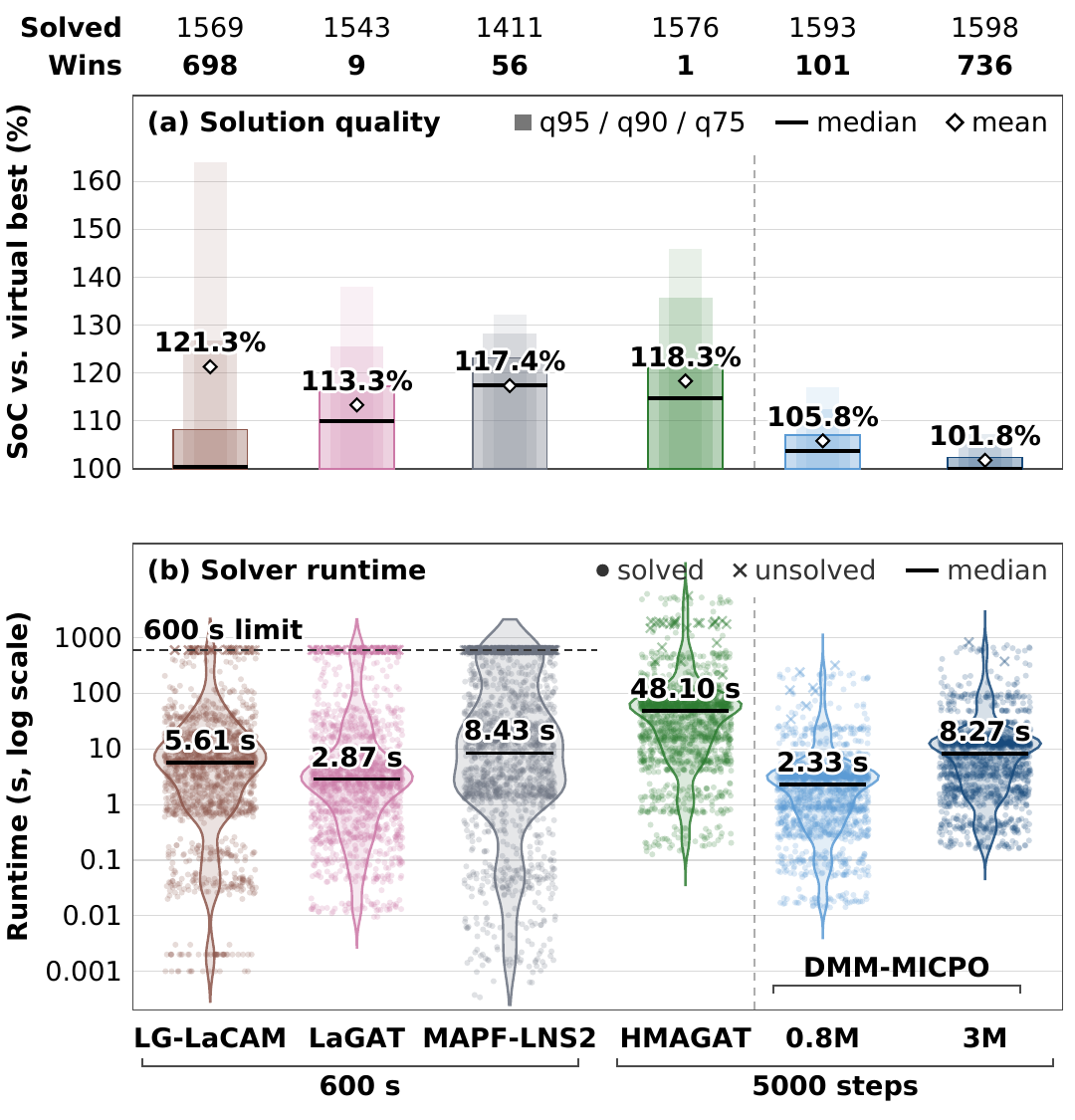}
    \caption{Results on 1,600 MovingAI tasks from 32 maps, each instantiated using all available start--goal pairs in its scenario file (i.e., the maximum available agent count).
    Annotations above the panels report the number of solved tasks (out of 1,600) and
    virtual-best SoC wins. The virtual best is the minimum SoC obtained by any
    of the six displayed methods on each task; ties count as wins for every
    tied method. \textbf{(a)} SoC relative to the virtual best on tasks solved
    by each method. Annotations show mean values, and the nested bands
    show the $q_{75}$, $q_{90}$, and $q_{95}$ quantiles. \textbf{(b)} Reported
    solver runtime on a logarithmic scale; dots and crosses denote all solved
    and unsolved tasks. Search-based and hybrid methods receive 600 seconds per
    task, while HMAGAT and DMM receive 5,000 environment steps. HMAGAT and DMM
    use CS--PIBT and RSE.}
    \label{fig:movingai1600}
\end{figure}

Second, the compact DMM-0.8M variant remains competitive despite its lighter
architecture. Across the evaluated domains and agent counts, it generally
performs comparably to or better than LC-MAPF-3M. This makes
DMM-0.8M a computationally lighter DMM configuration for the subsequent
large-scale evaluations.

Third, MICPO generally improves both DMM variants over their imitation-only
counterparts across the POGEMA domains. The MICPO-fine-tuned policies achieve
higher success rates, particularly at larger team sizes, while their SoC-ratio
distributions shift toward lower values. This pattern is observed for both the
0.8M- and 3M-parameter variants.

Comparing all methods, among solved instances, DMM-MICPO-3M has the
lowest median and narrowest interquartile range on every map type, with
per-domain medians ranging from 1.006 to 1.102, while MAGAT+ and HMAGAT show
the widest spreads and heaviest upper tails, with per-domain medians ranging
from 1.063 to 1.892. The lower medians and narrower interquartile ranges show that, among solved
instances, DMM-MICPO-3M typically remains closer to the LaCAM* reference cost
with less variation than MAGAT+ and HMAGAT. The wider upper tails of MAGAT+
and HMAGAT show that these methods also produce some much more
expensive solutions.

\begin{table*}[!t]
  \centering
    \caption{Results on matched $2304\times2304$ POGEMA mazes using four H100 GPUs.
    Each row aggregates four matched scenarios, one per maze layout.
    Final on goal is the final on-goal percentage; episodes stop when all agents
    are on goal or at $32\,768$ steps.
    Avg. episode length is the mean across the four scenarios, while Min. and Max.
    are the minimum and maximum episode lengths across them.
    Decision time is $10^6T/(NL)$ for total wall time $T$ (seconds, including
    prefill and overhead), $N$ agents and $L$ steps, counting all agents including
    skipped policy computations.
    Density is agents per free cell.
    Total time reports the same $T$ in minutes.
    Step, PIBT and decision times are mean $\pm$ SD across the four scenarios; all
    other values are means.}
  \label{tab:maze-1m-performance}
  \scriptsize
  \setlength{\tabcolsep}{2.8pt}
  \renewcommand{\arraystretch}{1.12}
  \begin{tabular*}{\textwidth}{
    @{\extracolsep{\fill}}
    lrcrrrrr
    @{\extracolsep{0pt}\,$\pm$\,}l
    @{\extracolsep{\fill}}r
    @{\extracolsep{0pt}\,$\pm$\,}l
    @{\extracolsep{\fill}}r
    @{\extracolsep{0pt}\,$\pm$\,}l
    @{\extracolsep{\fill}}rr
    @{}
  }
    \toprule
    & & & & \multicolumn{3}{c}{Episode length $\downarrow$}
    & \multicolumn{2}{c}{} & \multicolumn{2}{c}{}
    & \multicolumn{2}{c}{} & & \\
    \cmidrule(lr){5-7}
    Method
    & \smash{\shortstack{Agent\\count}}
    & \smash{\shortstack{Agent\\density (\%)}}
    & \smash{\shortstack{Final on goal\\(\%) $\uparrow$}}
    & Avg. & Min. & Max.
    & \multicolumn{2}{c}{\smash{\shortstack{Avg. step\\time (ms)}}}
    & \multicolumn{2}{c}{\smash{\shortstack{PIBT time\\(ms/step)}}}
    & \multicolumn{2}{c}{\smash{\shortstack{Decision time\\($\mu$s/agent/step)}}}
    & \smash{\shortstack{Peak GPU\\memory}}
    & \smash{\shortstack{Total time\\(min)}} \\
    \midrule

    DMM-MICPO-0.8M
    & \multirow{2}{*}{\small $1\,048\,576$}
    & \multirow{2}{*}{\small $27.86$}
    & $100.00$
    & $20\,867$ & $16\,345$ & $29\,508$
    & $763.1$ & $195.0$
    & $249.22$ & $62.86$
    & $0.745$ & $0.189$
    & $49.07$\,GiB
    & $256.9$ \\

    GPU-PIBT
    & & 
    & $89.15$
    & $32\,768$ & $32\,768$ & $32\,768$
    & $228.9$ & $6.7$
    & $186.26$ & $6.50$
    & $0.234$ & $0.006$
    & $34.01$\,GiB
    & $134.3$ \\

    \midrule

    DMM-MICPO-0.8M
    & \multirow{2}{*}{\small $524\,288$}
    & \multirow{2}{*}{\small $13.93$}
    & $100.00$
    & $8\,389$ & $7\,600$ & $8\,989$
    & $528.6$ & $40.0$
    & $22.02$ & $4.98$
    & $1.035$ & $0.077$
    & $28.77$\,GiB
    & $75.7$ \\

    GPU-PIBT
    & &
    & $95.46$
    & $32\,768$ & $32\,768$ & $32\,768$
    & $32.7$ & $1.2$
    & $10.56$ & $0.73$
    & $0.077$ & $0.002$
    & $17.84$\,GiB
    & $21.9$ \\

    \midrule

    DMM-MICPO-0.8M
    & \multirow{2}{*}{\small $262\,144$}
    & \multirow{2}{*}{\small $6.97$}
    & $100.00$
    & $5\,185$ & $4\,961$ & $5\,461$
    & $488.9$ & $11.6$
    & $2.73$ & $0.66$
    & $1.902$ & $0.046$
    & $18.63$\,GiB
    & $43.1$ \\

    GPU-PIBT
    & &
    & $97.83$
    & $32\,768$ & $32\,768$ & $32\,768$
    & $14.9$ & $0.5$
    & $1.89$ & $0.13$
    & $0.071$ & $0.002$
    & $9.76$\,GiB
    & $10.1$ \\

    \midrule

    DMM-MICPO-0.8M
    & \multirow{2}{*}{\small $131\,072$}
    & \multirow{2}{*}{\small $3.48$}
    & $100.00$
    & $4\,704$ & $4\,647$ & $4\,769$
    & $440.3$ & $4.2$
    & $1.16$ & $0.10$
    & $3.401$ & $0.032$
    & $11.32$\,GiB
    & $35.0$ \\

    GPU-PIBT
    & &
    & $98.91$
    & $32\,768$ & $32\,768$ & $32\,768$
    & $10.9$ & $0.1$
    & $1.21$ & $0.02$
    & $0.100$ & $0.002$
    & $5.71$\,GiB
    & $7.2$ \\

    \bottomrule
  \end{tabular*}
\end{table*}

\subsection{MovingAI Benchmark}
\label{sec:movingai-results}

Figure~\ref{fig:movingai1600} compares both DMM sizes with LG-LaCAM,
MAPF-LNS2, LaGAT, and HMAGAT. The clearest result is coverage. DMM-MICPO-3M solves 1,598 of the 1,600 tasks,
the highest coverage among all evaluated methods. DMM-MICPO-0.8M follows with
1,593 solved tasks, ahead of HMAGAT (1,576), LG-LaCAM (1,569), LaGAT (1,543),
and MAPF-LNS2 (1,411). Thus, both DMM configurations outperform the
search-based, hybrid, and learned baselines in coverage despite operating
reactively, without constructing a search tree or revisiting previously
executed decisions.

DMM also returns solutions with high quality. DMM-MICPO-3M matches the per-task
virtual-best SoC on 736 tasks, more than any other method, followed by LG-LaCAM
with 698. Its mean/median SoC ratios are 101.8/100.1\%, and its $q_{95}$ ratio is 107.3\%,
meaning that on 95\% of the instances its SoC is at most 7.3\% above the virtual best. 
The result is consistent across the benchmark: DMM-MICPO-3M solves
all 50 tasks on 31 of the 32 maps and 48 tasks on the remaining map, while its
map-wise mean virtual-best ratio never exceeds 108.9\%. This stability spans
maps with different topology, size, agent count, and agent density.

Among the baselines, LG-LaCAM provides the strongest combination of coverage
and typical-case solution quality. It solves 1,569 tasks, matches the virtual
best on 698, and has a median SoC ratio of 100.3\%. However, this performance is
not uniform across map families. LG-LaCAM exhibits a pronounced high-cost tail
on maze and room maps, as well as on game maps with narrow corridors, including
\texttt{den312d} and \texttt{lt\_gallowstemplar\_n}. Consequently, despite its
near-virtual-best median, its mean SoC ratio rises to 121.3\% and its $q_{95}$
ratio to 164.0\%.

\begin{figure*}[!thb]
    \centering
    \includegraphics[width=\textwidth]{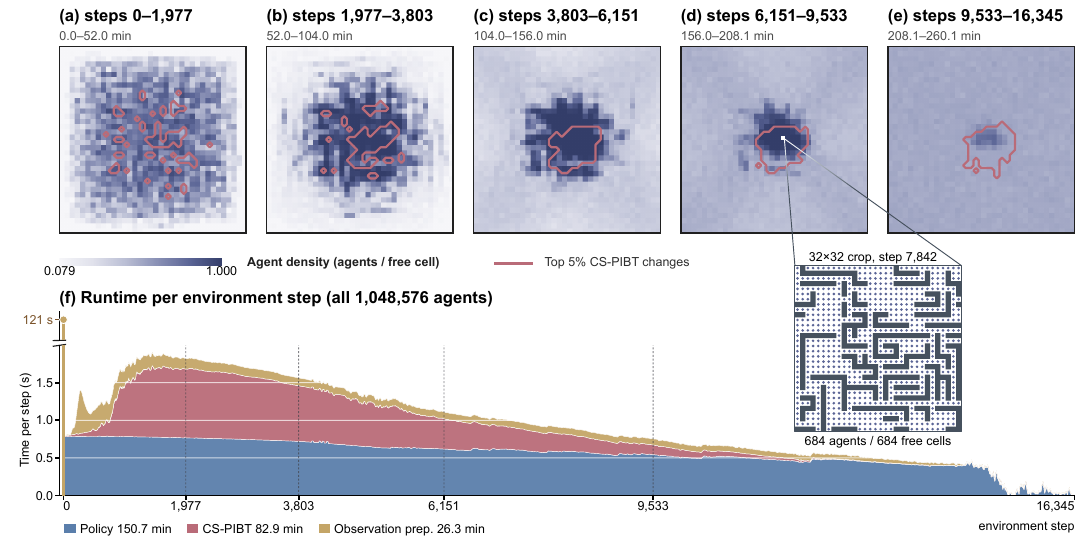}
    \caption{Congestion and computation during a successful
$1{,}048{,}576$-agent DMM-MICPO-0.8M rollout.
\textbf{(a--e)} Five consecutive windows of approximately 52 minutes of
measured step time. Purple shows mean agent presence per traversable cell
in each $64\times64$ region, including agents on goal; rose contours mark
the top 5\% of regions by CS--PIBT action changes within each window.
Central congestion clears as agents reach their goals.
\textbf{(f)} Per-step policy inference, shielding with action broadcast,
and observation construction times: the maximum across four GPUs for each
stage, smoothed over 32 steps. Faster steps let later windows cover more
steps. The broken axis marks the one-time 121-second cost-to-go prefill;
the legend gives episode totals by stage. All agents reach their goals
at step 16,345.}
    \label{fig:million-agent-rollout}
\end{figure*}

The smaller DMM configuration has the lowest median runtime: 2.33 seconds,
compared with 2.87 seconds for LaGAT, 5.61 seconds for
LG-LaCAM, 8.27 seconds for DMM-MICPO-3M, 8.43 seconds for MAPF-LNS2, and 48.10
seconds for HMAGAT. Under a 600-second end-to-end planning budget, the median
DMM-MICPO-0.8M and DMM-MICPO-3M runs would leave more than 99.6\% and 98.6\%
of the budget, respectively, for optional solution refinement such as LNS.
These runtime distributions describe the actual solver configurations and
stopping rules rather than an equalized inference budget: the search-based and
hybrid methods are time-limited, whereas HMAGAT and DMM are step-limited.

The two DMM configurations therefore define complementary operating points:
DMM-MICPO-0.8M prioritizes runtime while retaining the second-highest coverage,
whereas DMM-MICPO-3M provides the best coverage and solution quality at a
runtime comparable to MAPF-LNS2. These results demonstrate that a
learning-based reactive planner can compete with state-of-the-art MAPF solvers,
including search-based, hybrid, and learned approaches, while providing high
coverage, fast valid solutions, and consistently strong solution quality.

\subsection{Scaling DMM to One Million Agents}
\label{sec:million-agents}

Table~\ref{tab:maze-1m-performance} compares completion and computational
performance. DMM solves all 16 instances, whereas GPU-PIBT reaches the
horizon in every run. As agent density increases from 3.48\% to 27.86\%,
GPU-PIBT's mean final on-goal fraction falls from 98.91\% to 89.15\%.
DMM's mean episode length increases from 4,704 to 20,867 steps, with
all million-agent scenarios solved in 16,345--29,508 steps.
GPU-PIBT's per-step runtime grows sharply with density: an $8\times$
increase in agent count raises mean step time approximately $21\times$,
from 10.9 to 228.9 milliseconds. This is consistent with larger conflict
groups limiting parallelism at higher densities.

DMM's mean per-step cost grows more slowly than the population:
an $8\times$ increase in agent count raises mean step time only
$1.73\times$, from 440.3 to 763.1 milliseconds. Peak memory increases
from 11.32 to 49.07 GiB per GPU as more agents require more storage
for cached BFS distances and agent state. Longer episodes nevertheless
raise mean total runtime from 35.0 minutes to 4.28 hours. Within each
rollout, inference skipping reduces policy computation as agents reach
their goals. GPU-PIBT runs faster and uses less memory, but its runtimes
correspond to incomplete, horizon-limited runs.

Figure~\ref{fig:million-agent-rollout} examines a million-agent maze
run in detail. All 1,048,576 agents reach their goals after 16,345 steps,
with a total runtime of 4.40 hours. Five equal-duration windows of
approximately 52 minutes of measured step time show how congestion
and computation change throughout the rollout.

\begin{table}[!b]
\centering
\caption{Sample frequency over the four joint actions at the shared ambiguous
corridor state, as percentages (1,000 samples per seed, 5 training seeds;
mean $\pm$ 95\% Student-$t$ confidence interval across seeds). For DMM,
$\beta$ denotes the shared teacher-forcing floor, $\beta_0=\beta_r$. Joint
actions are abbreviated as WL = (wait, left), RW = (right, wait),
WW = (wait, wait), and RL = (right, left).}
\label{tab:corridor_freq}
\footnotesize
\setlength{\tabcolsep}{2pt}
\renewcommand{\arraystretch}{1.2}
\newcommand{\corridorvalue}[2]{#1\,{\tiny$\pm$#2}}
\begin{tabular*}{\columnwidth}{@{\extracolsep{\fill}}lcccc@{}}
\toprule
& \multicolumn{2}{c}{Valid} & \multicolumn{2}{c}{Invalid} \\
\cmidrule(lr){2-3}\cmidrule(l){4-5}
Method & WL (\%) & RW (\%) & WW (\%) & RL (\%) \\
\midrule
DMM ($\beta=0.8$)
    & \cellcolor{best}\corridorvalue{51.4}{9.0}
    & \cellcolor{best}\corridorvalue{43.4}{9.4}
    & \cellcolor{best}\corridorvalue{2.2}{0.5}
    & \cellcolor{best}\corridorvalue{3.0}{1.0} \\
DMM ($\beta=0$)
    & \corridorvalue{24.5}{0.4}
    & \corridorvalue{25.6}{0.5}
    & \corridorvalue{23.3}{0.2}
    & \corridorvalue{26.6}{0.2} \\
\midrule
LC-MAPF
    & \corridorvalue{25.7}{0.1}
    & \corridorvalue{24.2}{0.2}
    & \corridorvalue{22.7}{0.4}
    & \corridorvalue{27.4}{0.2} \\
MAGAT+
    & \corridorvalue{24.6}{0.3}
    & \corridorvalue{24.9}{0.4}
    & \corridorvalue{23.5}{0.4}
    & \corridorvalue{26.9}{0.2} \\
HMAGAT
    & \corridorvalue{25.6}{1.4}
    & \corridorvalue{24.7}{2.0}
    & \corridorvalue{22.5}{2.2}
    & \corridorvalue{27.1}{1.4} \\
\bottomrule
\end{tabular*}
\end{table}

Traffic concentrates around the map center before clearing as agents
reach their goals. Regions where CS--PIBT most often changes proposed
actions overlap with this central congestion, showing where collision
resolution places the greatest demands on the policy's proposed moves.
As congestion clears and inference skipping reduces policy computation,
steps become faster. Shielding nevertheless remains a substantial
runtime cost despite GPU parallelism.

\subsection{Corridor Conflict}
\label{sec:corridor}

DMM with the retained teacher-forcing floor $\beta_0=\beta_r=0.8$ places
94.8\% of its samples on the two valid resolutions (Table~\ref{tab:corridor_freq}). In contrast, LC-MAPF, MAGAT+, and
HMAGAT place approximately equal probability on all four joint actions, with
about 50\% of their samples falling on the two valid resolutions. This is
consistent with the factorization-gap analysis in
Eq.~\eqref{eq:decentralized-factorization-gap}: the individual actions remain
plausible, but independent sampling recombines them into invalid joint
outcomes. At the ambiguous state, the corridor reduces to an
anti-coordination game with two optimal joint actions, a structure known to be
difficult for independently acting agents in cooperative multi-agent
learning~\citep{claus1998dynamics,matignon2012independent}.
DMM instead produces a strongly correlated joint-action distribution
by allowing agents' evolving intents to influence subsequent votes before
commitment. Although DMM is trained with $K=4$, evaluating the same checkpoints
with 8 and 12 refinement rounds further increases the valid-action frequency
(Figure~\ref{fig:corridor}); the full test-time-depth and teacher-forcing ablations are reported in
\appref{sec:corridor-ablation}.

With a zero teacher-forcing floor, DMM loses its concentration on the two
valid resolutions and approaches the near-uniform frequencies of
LC-MAPF, MAGAT+, and HMAGAT. Thus, a nonzero teacher-forcing floor
is important for learning the correlated refinement behavior.

\section{Conclusion}

In this work, we identified a limitation of decentralized MAPF policies that
sample agents' final actions independently: individually plausible choices can
still combine into an incompatible joint action. DMM addresses this by
refining stochastic action intents through local communication before
simultaneous commitment, allowing agents' evolving choices to influence one
another while preserving decentralized execution. We further introduced MICPO,
a critic-free reinforcement-learning method that fine-tunes this multi-round
decision process from shared task-level outcomes.

Across POGEMA, DMM generally achieves higher success rates and lower solution
costs than the evaluated learned policies, while MICPO further improves both
DMM variants over their imitation-pretrained counterparts. At the solver level,
DMM-MICPO-3M solves 1,598 of 1,600 MovingAI tasks, achieving the highest
coverage among all evaluated methods. The compact DMM-MICPO-0.8M further
demonstrates scalability, solving every tested large-scale instance up to
1,048,576 simultaneously acting agents.

A remaining limitation is that DMM does not enforce joint-action feasibility
as a hard constraint. Although MICPO penalizes blocked action proposals through
the task-level objective, incompatible intermediate choices are not explicitly
excluded during refinement. Collision-free execution may therefore still
require an external mechanism such as CS--PIBT, which makes the system centralized in our MovingAI and large-scale experiments. Incorporating explicit
feasibility constraints into the refinement process is a natural
direction for future work.

Overall, these results show that coupling agents' evolving action choices
before commitment can improve decentralized MAPF beyond independent
one-shot action sampling.

\bibliographystyle{unsrtnat}
\bibliography{references}

\setcounter{section}{0}
\renewcommand{\thesection}{\Alph{section}}

\section*{Appendix Contents}

\begin{center}
\begingroup
\small
\renewcommand{\arraystretch}{1.15}
\setlength{\tabcolsep}{4pt}

\begin{tabularx}{\columnwidth}{@{}lX@{}}
\toprule
\textbf{Section} & \textbf{Contents} \\
\midrule

\appref{sec:factorization-gap} &
Proof of the decentralized factorization gap \\

\addlinespace[2pt]
\appref{sec:dmm-architectures} &
DMM-3M and DMM-0.8M architecture variants and implementation details \\

\addlinespace[2pt]
\appref{sec:micpo-details} &
MICPO scenario generation, rollouts, trajectory filtering,
timestep sampling, and multi-round replay \\

\addlinespace[2pt]
\appref{sec:hyper} &
Imitation-pretraining and MICPO fine-tuning hyperparameters \\

\addlinespace[2pt]
\appref{sec:movingai-extended} &
MovingAI benchmark composition, solver budgets, evaluation metrics,
and Repeat-State Escape \\

\addlinespace[2pt]
\appref{sec:gpu-pipeline} &
GPU-accelerated environment dynamics, observation construction,
communication, and memory-bounded execution \\

\addlinespace[2pt]
\appref{sec:round-ablation} &
Inference-time refinement-depth ablation across POGEMA domains
before and after MICPO fine-tuning \\

\addlinespace[2pt]
\appref{sec:z-ablation} &
Intent-state communication ablations of message components,
intent evolution, and message assignment \\

\addlinespace[2pt]
\appref{sec:corridor-ablation} &
Corridor ablations of test-time refinement depth, teacher-forcing
mechanisms, floor values, and training schedules \\

\addlinespace[2pt]
\appref{sec:movingai-rse-comparison} &
MovingAI reactive-policy results with and without Repeat-State Escape \\

\bottomrule
\end{tabularx}
\endgroup
\end{center}

\section{Decentralized Factorization Gap Proof}
\label{sec:factorization-gap}

We use the notation introduced in the factorization-gap analysis in the main
text; all expectations, entropies, and mutual informations are computed under
$P^\star$.
For any product executor $Q_q$, the expected joint KL divergence \mbox{decomposes as}
\begin{align}
&\mathbb{E}_{X}
D_{\mathrm{KL}}\!\left(
P^\star(A\mid X)
\,\middle\|\,
\prod_{u\in U} q_u(A_u\mid X_u)
\right)
\nonumber\\
&\quad =
\underbrace{\mathrm{TC}(A\mid X)}
_{\text{action coupling}}
+
\underbrace{
\sum_{u\in U} I(A_u;X_{-u}\mid X_u)
}_{\text{missing local information}}
\nonumber\\
&\qquad\quad +
\underbrace{
\sum_{u\in U}
\mathbb{E}_{X_u}
D_{\mathrm{KL}}\!\left(
P^\star(A_u\mid X_u)
\,\middle\|\,
q_u(A_u\mid X_u)
\right)
}_{\text{learning error}}.
\label{eq:factorization-gap-decomposition}
\end{align}

To prove this identity, expand the product form of $Q_q$:
\begin{align*}
&\mathbb{E}_{X}
D_{\mathrm{KL}}\!\left(
P^\star(A\mid X)
\,\middle\|\,
\prod_{u\in U} q_u(A_u\mid X_u)
\right) \\
&\qquad =
-H(A\mid X)
+
\sum_{u\in U}
\mathbb{E}\!\left[-\log q_u(A_u\mid X_u)\right].
\end{align*}
Each summand is a local cross-entropy and therefore satisfies
\begin{align*}
\mathbb{E}\!\left[-\log q_u(A_u\mid X_u)\right]
&= H(A_u\mid X_u) \\
&\hspace{-2.5em}
+\mathbb{E}_{X_u}
D_{\mathrm{KL}}\!\left(
P^\star(A_u\mid X_u)
\,\middle\|\,
q_u(A_u\mid X_u)
\right).
\end{align*}
Next, because $X=(X_u,X_{-u})$,
\[
    H(A_u\mid X_u)
    =
    H(A_u\mid X)
    +
    I(A_u;X_{-u}\mid X_u).
\]
Substituting this identity into the expanded KL and collecting entropy terms
gives Eq.~\eqref{eq:factorization-gap-decomposition}, since
\[
    \sum_{u\in U}H(A_u\mid X)-H(A\mid X)
    =
    \mathrm{TC}(A\mid X).
\]

All three terms in Eq.~\eqref{eq:factorization-gap-decomposition} are
nonnegative. With sufficient capacity and ideal cross-entropy optimization,
each local policy converges to its Bayes-optimal conditional,
\[
    q_u^\star(a_u\mid x_u)
    =
    P^\star(A_u=a_u\mid X_u=x_u),
\]
and the learning-error terms vanish. The remaining irreducible error is
therefore
\[
    \mathrm{TC}(A\mid X)
    +
    \sum_{u\in U}I(A_u;X_{-u}\mid X_u).
\]
This is the minimum in Eq.~\eqref{eq:decentralized-factorization-gap}.

\section{DMM Architecture Variants}
\label{sec:dmm-architectures}

\begin{figure*}[t]
    \centering
    \includegraphics[width=\textwidth]{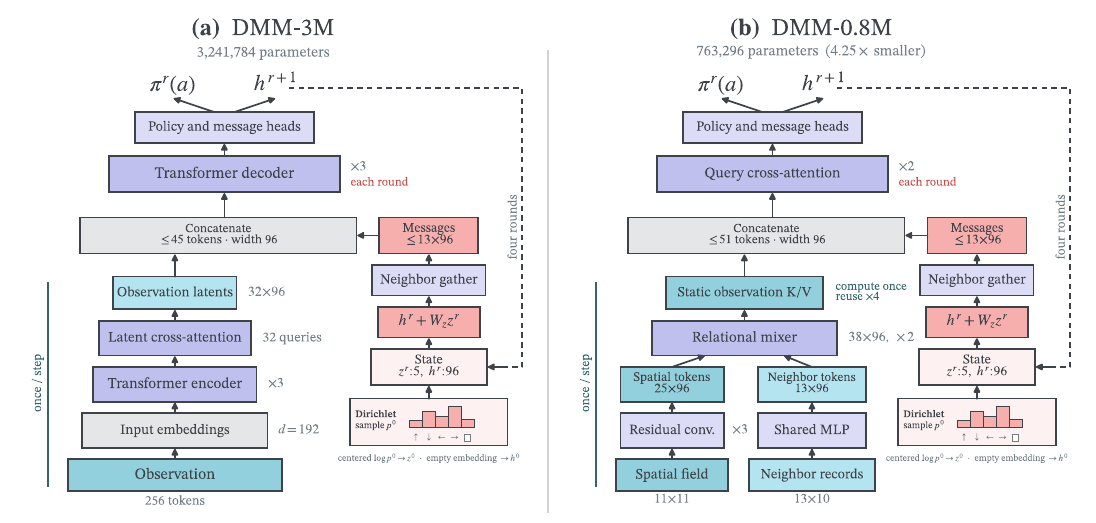}
    \caption{
    DMM architecture variants.
    \textbf{(a)} DMM-3M retains the Transformer encoder--decoder backbone of
    LC-MAPF. The tokenized observation is compressed into 32 latent tokens,
    which are combined with the current neighbor messages and processed by the
    decoder at every refinement round.
    \textbf{(b)} DMM-0.8M uses a structured observation encoder that separately
    processes the spatial field and local agent records. The resulting
    observation tokens are mixed once per environment step, and their
    cross-attention key/value projections are reused across refinement rounds;
    only the current messages and query state change between rounds.
    Both architectures implement the same DMM intent-refinement process.
    }
    \label{fig:dmm-variants}
\end{figure*}

Section~\ref{sec:method} defines the DMM intent-refinement mechanism at the
policy level. We instantiate this mechanism with two network architectures,
DMM-3M and DMM-0.8M, shown in Figure~\ref{fig:dmm-variants}. Both use the
same action-intent representation, intent-augmented messages, local
communication graph, and iterative voting procedure. They differ in how the
local observation is encoded and how the resulting observation representation
is combined with the changing messages at each refinement round.

\paragraph{DMM-3M.}
The DMM-3M architecture is shown in Figure~\ref{fig:dmm-variants}(a). DMM-3M retains the observation encoder and communication-decoder structure of
LC-MAPF~\citep{lcmapf2026}. Each agent receives the 256-token observation
described in Section~\ref{sec:experimental-setup}. Tokens are embedded at width 192 and
processed by three Transformer~\citep{vaswani2017attention} encoder blocks.
A set of 32 learned latent queries then cross-attends to the encoded
observation~\citep{jaegle2021perceiver}, producing a fixed $32\times96$
observation representation for the current environment step.

During each refinement round, the intent-augmented messages of up to 13 local
agents are gathered and concatenated with the 32 observation latents. The
resulting sequence, containing at most 45 tokens of width 96, is processed by
three Transformer decoder blocks. A learned action/message query then
cross-attends to this sequence to produce the per-round feature used by the
policy and message heads. These heads output the five action logits
$\phi_u^r$ and the next message feature $h_u^r$, respectively. The observation
latents remain unchanged across refinement rounds; only the intent and
communication state evolve. DMM-3M contains 3,241,784 trainable parameters.

\paragraph{DMM-0.8M.}
The DMM-0.8M architecture is shown in Figure~\ref{fig:dmm-variants}(b). DMM-0.8M preserves the same DMM refinement interface but replaces the
Transformer observation backbone and full per-round decoder with a structured,
lighter-weight architecture. It uses the same 256-token observation layout as
DMM-3M, whose two informative components are the $11\times11$ spatial field
and 13 local agent records of 10 tokens each.

The 121 spatial tokens are first embedded at width 96 and arranged as an
$11\times11$ feature map. A convolutional stem followed by three residual
convolutional blocks processes this map, after which adaptive pooling produces
a $5\times5$ representation, or 25 spatial tokens. In parallel, each
10-token agent record is embedded and processed by a shared MLP, producing
13 agent tokens of width 96. Learned spatial-position, agent-slot, and
token-type embeddings distinguish the two sources. The 25 spatial and
13 agent tokens are then concatenated and processed by two relational
self-attention blocks, yielding a 38-token observation representation.

This representation is computed once per environment step. To reduce the
computation repeated during intent refinement, DMM-0.8M uses two
query cross-attention blocks in place of the DMM-3M decoder. For each block,
the key and value projections of the 38 observation tokens are precomputed
once and reused across all refinement rounds. At round $r$, the current
intent-augmented messages are gathered from up to 13 local agents and
projected dynamically. A learned query attends jointly to the static
observation representation and these round-specific messages, and the updated
query is passed through the two cross-attention blocks sequentially. The final
96-dimensional query feature is mapped by the same type of policy and message
heads to the action logits $\phi_u^r$ and next message feature $h_u^r$.

DMM-0.8M therefore moves most observation-dependent computation outside the
refinement loop while retaining fresh message-dependent computation at every
round. It contains 763,296 trainable parameters, making it about
$4.25\times$ smaller than DMM-3M while leaving the intent update,
communication semantics, and final action rule unchanged.

\section{MICPO Training Procedure}
\label{sec:micpo-details}

Section~\ref{sec:method} defines the MICPO objective, matched rollout groups,
and bounded replay. Here we provide the implementation details of scenario
generation, trajectory filtering, timestep sampling, and multi-round replay
(Algorithm~\ref{alg:micpo}).

\paragraph{Training scenario generation.}
Training scenarios are generated online. For each scenario, the map geometry
and start--goal configuration are generated from independent random seeds.
Training uses a mixture of maze and randomly obstructed maps, with maze
sampling probability $p_{\mathrm{maze}}$. Randomly obstructed maps use obstacle
density $p_{\mathrm{obs}}$; maps with disconnected free-space regions are
rejected and resampled before agent starts and goals are generated.

The number of agents $N$ is fixed within a training run. Map height
$H_{\mathrm{map}}$ and width $W_{\mathrm{map}}$ are sampled independently from
the ranges
$[H_{\mathrm{map}}^{\min},H_{\mathrm{map}}^{\max}]$ and
$[W_{\mathrm{map}}^{\min},W_{\mathrm{map}}^{\max}]$, respectively.
Varying the map dimensions exposes the policy to different agent densities
and congestion levels while keeping $N$ fixed. Start and goal locations are
sampled on the generated map using POGEMA.

\paragraph{Rollout collection.}
At each environment timestep, DMM executes all $K$ refinement rounds before
committing an environment action. A trajectory ends when all agents
simultaneously occupy their goals or when the maximum rollout horizon
$T_{\max}$ is reached. Agents are not absorbed at their goals and may move
away again on subsequent timesteps.

For policy replay, the implementation retains the observation, communication
neighborhood, sampled initial intent $z_t^0$, the sequence of refinement votes
$y_t^{1:K}$, and the corresponding log-probabilities under
$\pi_{\mathrm{old}}$ for each \mbox{environment timestep}.

\begin{algorithm}[!t]
\caption{MICPO procedure}
\label{alg:micpo}
\begin{algorithmic}[1]

\STATE Initialize $\pi_\theta$ from imitation pretraining
\STATE Set $\pi_{\mathrm{ref}}\leftarrow\pi_\theta$ and keep it frozen

\FOR{each MICPO iteration}
    \STATE Set $\pi_{\mathrm{old}}\leftarrow\pi_\theta$
    \STATE Sample $B$ training scenarios

    \FOR{each sampled scenario}
        \FOR{$m=1,\ldots,M$}
            \STATE Initialize a rollout group of $G$ trajectories
                   $\zeta_1,\ldots,\zeta_G$ from the same scenario
            \FOR{each rollout timestep $t$}
                \STATE Sample an initial intent state $z_t^0$ and share it across
                       the $G$ trajectories
                \STATE Sample refinement votes independently across trajectories
                       for $K$ rounds under $\pi_{\mathrm{old}}$
                \STATE Store the sampled votes and their log-probabilities under
                       the corresponding $p_{\mathrm{old}}$
            \ENDFOR
        \ENDFOR
    \ENDFOR

    \FOR{each rollout group}
        \STATE Compute $R(\zeta_g)$ and $\hat A(\zeta_g)$
               for $g=1,\ldots,G$
        \STATE Let $\mathcal I$ contain the indices of the
               $\kappa$ lowest- and $\kappa$ highest-return trajectories
        \FOR{$g\in\mathcal I$}
            \STATE Sample $S$ valid timesteps from $\zeta_g$
        \ENDFOR
    \ENDFOR

    \FOR{each replay minibatch}
        \STATE Replay all $K$ refinement rounds from the stored initial intent
               $z_t^0$ and vote sequence $y_{g,t}^{1:K}$
        \STATE Recompute message features $h_{u,g,t}^r$ and action distributions
               $p_{\theta,u,g,t}^r$ and $p_{\mathrm{ref},u,g,t}^r$
               for all agents $u$ and rounds $r=1,\ldots,K$
        \STATE Update $\theta$ by minimizing
               $\mathcal{L}=\mathcal{L}_{\mathrm{clip}}+\mathcal{L}_{\mathrm{KL}}$
    \ENDFOR
\ENDFOR

\end{algorithmic}
\end{algorithm}

\paragraph{Trajectory filtering and timestep sampling.}
For each matched group, $R(\zeta)$ and $\hat A(\zeta)$ are first computed for
all $G$ collected trajectories as defined in Section~\ref{sec:method}.
The trajectories are then ranked by $R(\zeta)$, and the $\kappa$ trajectories
with the lowest returns together with the $\kappa$ trajectories with the
highest returns are retained. The retained trajectories use the advantages
computed from all $G$ trajectories.

For each retained trajectory $\zeta$, $S$ valid environment timesteps are
sampled uniformly. Sampling is performed without replacement when the
trajectory contains at least $S$ valid timesteps and with replacement
otherwise. Timestep sampling is applied only along the environment-time
dimension: selecting timestep $t$ retains all $K$ refinement rounds associated
with that decision.

\paragraph{Multi-round replay.}
Intermediate message features are not stored in the replay data. For a
selected timestep, replay begins from the stored observation, neighborhood
information, and initial intent $z_t^0$. The stored vote sequence
$y_t^{1:K}$ determines the successive intent updates, while the policy is
evaluated sequentially through all $K$ rounds to recompute the corresponding
message features and action distributions.

The stored old-policy log-probabilities are used in the round-level importance
ratios defined in Section~\ref{sec:method}. The same replay sequence is
evaluated under the frozen reference policy to compute the categorical KL
terms. At the beginning of each MICPO iteration,
$\pi_{\mathrm{old}}$ is synchronized with the current policy
$\pi_\theta$, whereas $\pi_{\mathrm{ref}}$ remains fixed at the
imitation-pretrained checkpoint.

\section{Training Hyperparameters}
\label{sec:hyper}

Table~\ref{tab:hyperparameters} summarizes the imitation-pretraining and
MICPO configurations used for DMM-3M and DMM-0.8M. Architectural details are
given in \appref{sec:dmm-architectures}, and the MICPO rollout and replay
procedure is described in \appref{sec:micpo-details}.

\begin{table}[t]
\centering
\small
\caption{Model and training hyperparameters for DMM variants.}
\label{tab:hyperparameters}
\setlength{\tabcolsep}{3.5pt}
\begin{tabular}{@{}lcc@{}}
\toprule
\textbf{Parameter}
& \textbf{DMM-3M}
& \textbf{DMM-0.8M} \\
\midrule

\multicolumn{3}{@{}l}{\textit{Imitation pretraining}} \\

Effective batch size
    & 512
    & 800 \\

Training iterations
    & $1{,}000{,}000$
    & $1{,}000{,}000$ \\

Maximum learning rate
    & $6\times10^{-4}$
    & $6\times10^{-4}$ \\

Minimum learning rate
    & $6\times10^{-5}$
    & $6\times10^{-5}$ \\

Warm-up iterations
    & $2{,}000$
    & $2{,}000$ \\

Intent-update step size ($\delta$)
    & 0.25
    & 0.25 \\

Log-smoothing constant ($\eta$)
    & $1\times10^{-8}$
    & $1\times10^{-8}$ \\

Group-normalization threshold ($\tau$)
    & $1\times10^{-6}$
    & $1\times10^{-6}$ \\

Initial-intent TF start / floor ($\beta_0$)
    & 1.0 / 0.8
    & 1.0 / 0.8 \\

Round TF start / floor ($\beta_r$)
    & 1.0 / 0.8
    & 1.0 / 0.8 \\

\midrule
\multicolumn{3}{@{}l}{\textit{MICPO fine-tuning}} \\

Outer iterations
    & 500
    & 500 \\

Training agents ($N$)
    & 32
    & 32 \\

Map height / width range
    & 9--13
    & 9--13 \\

Maze probability ($p_{\mathrm{maze}}$)
    & 0.5
    & 0.5 \\

Random-map obstacle density ($p_{\mathrm{obs}}$)
    & 0.2
    & 0.2 \\

Maximum rollout horizon ($T_{\max}$)
    & 128
    & 128 \\

Scenarios per iteration ($B$)
    & 4
    & 4 \\

Groups per scenario ($M$)
    & 3
    & 3 \\

Trajectories per group ($G$)
    & 24
    & 24 \\

Trajectories retained per extreme ($\kappa$)
    & 4
    & 4 \\

Timesteps per retained trajectory ($S$)
    & 16
    & 16 \\

Clip parameter ($\varepsilon$)
    & 0.2
    & 0.2 \\

KL coefficient ($\alpha_{\mathrm{KL}}$)
    & 0.01
    & 0.01 \\

Off-goal weight ($w_c$)
    & 1.0
    & 1.0 \\

Blocked-action weight ($w_b$)
    & 0.3
    & 0.3 \\

Learning rate
    & $1\times10^{-6}$
    & $1\times10^{-6}$ \\

\bottomrule
\end{tabular}
\end{table}

\section{MovingAI Benchmark Protocol}
\label{sec:movingai-extended}

\paragraph{Benchmark composition.}
Each of the 33 MovingAI maps provides 25 even and 25 random scenarios. We use
the maximum-agent instance from every scenario, with populations ranging from
32 to 8,000 agents. We omit only \texttt{maze-128-128-1}, for which preliminary
runs found that no evaluated method solved any maximum-agent task. This
leaves 32 maps and 1,600 tasks. An unsolved task counts as a failure and
cannot contribute a solution-quality win.

\paragraph{Search-based and hybrid solvers.}
LG-LaCAM and MAPF-LNS2 represent two complementary search paradigms, while
LaGAT combines a learned MAGAT+ policy with LaCAM's search structure. Because
these methods can continue constructing and exploring alternatives, each run
is limited to 600 wall-clock seconds. We request the first valid solution from
LG-LaCAM and disable its optional iterative post-solution improvement. The
reported runtime is time to the returned solution, or the full budget for an
unsolved task.

\paragraph{Reactive learned solvers.}
HMAGAT, DMM-MICPO-0.8M, and DMM-MICPO-3M act directly in the environment and are
limited to 5,000 environment steps. Their action proposals are processed by
the same CS--PIBT collision shield. In the main comparison, all three also use
Repeat-State Escape (RSE) to prevent a reactive rollout from cycling among
previously visited joint configurations. Results without RSE are reported in
\appref{sec:movingai-rse-comparison}.

RSE stores the complete configuration history in a hash set. When CS--PIBT
produces a duplicate configuration, RSE considers agents in PIBT priority
order and selects the first unfinished agent whose proposed transition can be
forbidden without removing all available actions. The prohibition is local to
the current timestep, and CS--PIBT is invoked again from the same state to
construct a different successor. Constraints accumulate only during this
retry process and are discarded after an action is executed. RSE therefore
maintains no search tree, open list, rollback operation, or persistent search
node; it transfers LaCAM's repeated-state avoidance principle to a purely
reactive solver.

\paragraph{Reported quantities.}
We report the number of solved tasks, per-task virtual-best SoC wins, the
distribution of SoC relative to the virtual best on tasks solved by each
method, and wall-clock runtime. The virtual best is the minimum SoC obtained
by any of the evaluated methods on a task, with ties counted as wins for every
tied method. Since the search-based and hybrid solvers are time-limited whereas
the reactive policies are step-limited, runtime reflects the solver
configuration rather than an equalized compute budget.

\section{GPU-Accelerated Environment and Observation Pipeline}
\label{sec:gpu-pipeline}

The original POGEMA implementation generated observations and executed environment transitions on the CPU. While efficient for moderate-scale runs, this setup became a throughput bottleneck when scaling to many agents or running large batches of parallel rollouts, as each step required transferring data between CPU and GPU and limited overall simulation speed.

To overcome this, we reimplement both the environment dynamics and the observation pipeline to run entirely on the GPU. The full environment state (agent positions, goals, obstacle grid, and collision-resolution buffers) is kept in VRAM. At each timestep, a pipeline of CUDA kernels processes all agent actions: candidate moves are proposed, swap conflicts cancelled, and vertex collisions resolved through an iterative cascade using atomic priority arbitration, following the same soft-collision semantics as the original environment. No environment or observation tensors are moved back to the host during the rollout loop.

The observation construction is also rewritten for the GPU. Egocentric cost-to-go maps are obtained from a batched breadth-first search where each agent is handled by a single GPU block using shared memory; the search expands from the agent's goal outward and stops once the local observation window centered on the agent's current position is filled, avoiding full-map traversal. Rather than recomputing distances from scratch at every step, we cache larger raw distance windows. Agents whose Chebyshev distance from the cache center exceeds a margin recompute the BFS; all others extract their observation patch via a dedicated kernel that crops and normalizes the cached distances in one pass, deriving next-action tokens without a new BFS. Neighbor information (relative positions, goal offsets, and action histories) is assembled on-device by a batched CUDA kernel: for each agent we identify visible neighbors within a fixed Chebyshev radius, rank them by Manhattan distance, and encode the features. All observation tokens are concatenated and padded to a fixed context length, yielding a batched tensor in VRAM that feeds directly into the policy network, never leaving the device. The same local neighborhood also defines the sparse communication graph (top-$k$ neighbor indices) used by the policy.

To handle very large agent counts without exhausting GPU memory, we employ two complementary chunking strategies. BFS chunking processes cache-miss agents (or all agents when caching is disabled) in smaller sub-batches, limiting peak allocations for frontier and visited grids. Agent chunking partitions agents inside the policy encoder and, within each communication round, across decoder receivers, bounding the size of activations after a synchronization point that assembles the full per-agent message table. Communication itself is already sparse: each agent attends only to this precomputed top-$k$ set rather than to all other agents; chunking does not alter this graph, nor the computation. Both strategies are semantically equivalent to the monolithic path.

\section{Inference-Time Refinement Depth}
\label{sec:round-ablation}

\begin{figure*}[t]
\centering
\includegraphics[width=1.0\textwidth,trim=0 120 0 0,clip]{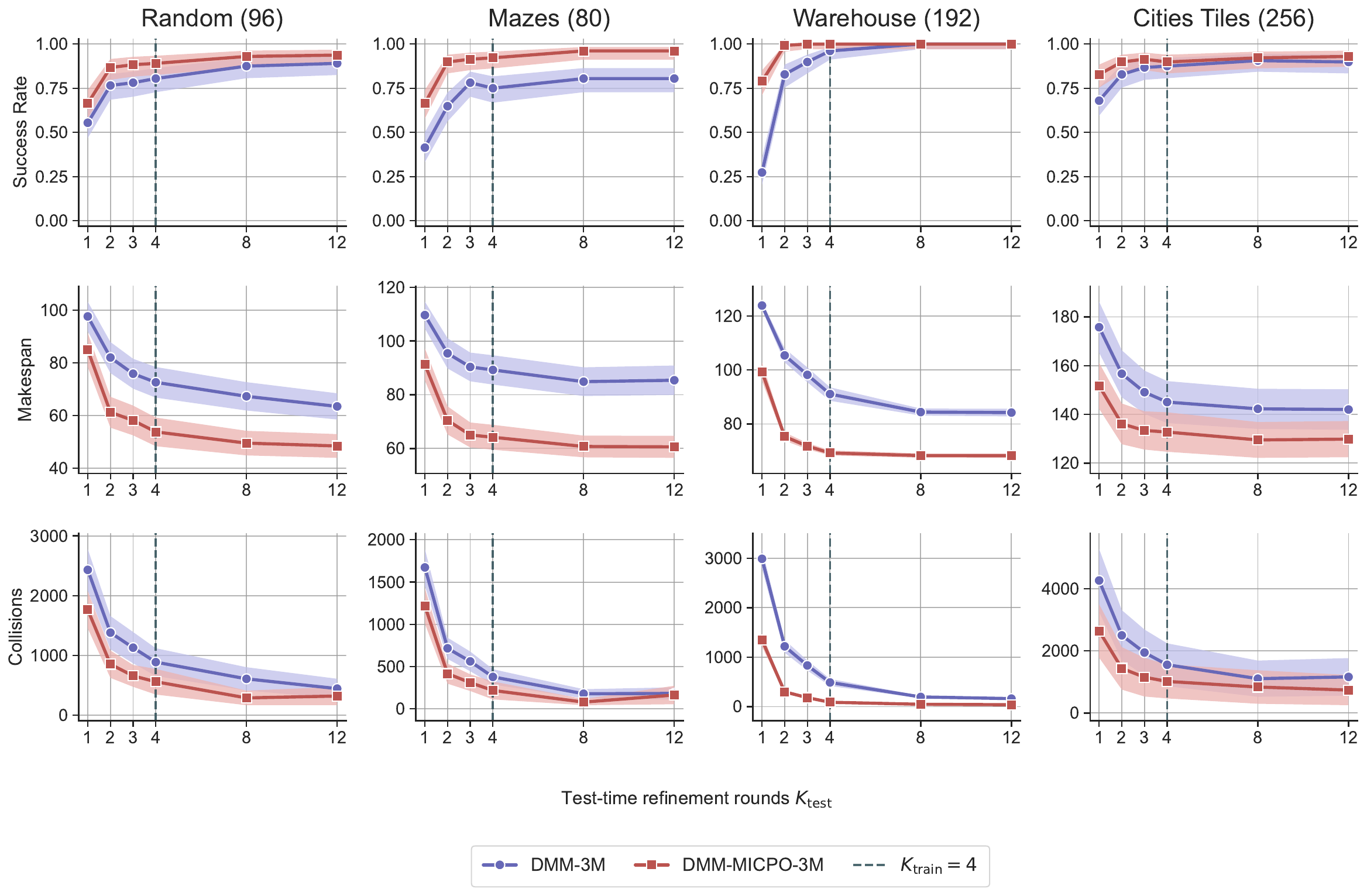}\\[2pt]
$\vcenter{\hbox{\includegraphics[width=0.1958\textwidth,trim=455 69 435 645,clip]{figures/07_refinement_depth_ablation.pdf}}}$\hspace{3em}%
$\vcenter{\hbox{\includegraphics[width=0.3653\textwidth,trim=383 4 307 690,clip]{figures/07_refinement_depth_ablation.pdf}}}$
\caption{Effect of inference-time refinement depth at the largest evaluated
team size in each POGEMA domain. DMM-3M and DMM-MICPO-3M are both trained
with $K_{\mathrm{train}}=4$ refinement rounds; at evaluation, the same fixed
checkpoints are executed with
$K_{\mathrm{test}}\in\{1,2,3,4,8,12\}$ without retraining.
Rows report success rate, makespan, and agent-agent collisions, respectively.
Shaded regions show 95\% confidence intervals across $n=128$ validation
instances per domain, with the same instances used across refinement depths
and policies. Wilson score intervals are used for success rate, while makespan and collision
counts use normal intervals of the form
$\bar{x}\pm1.96\,s/\sqrt{n}$. The
vertical dashed line marks the training depth $K_{\mathrm{train}}=4$.}
\label{fig:round-ablation}
\end{figure*}

DMM is trained with $K_{\mathrm{train}}=4$ refinement rounds. To test whether
the learned refinement process is tied to this depth, we keep the policy
parameters fixed and vary only the number of rounds executed at inference,
$K_{\mathrm{test}}\in\{1,2,3,4,8,12\}$. We evaluate both DMM-3M before
MICPO fine-tuning and DMM-MICPO-3M after fine-tuning, using the same
evaluation protocol as in the main POGEMA benchmark. Figure~\ref{fig:round-ablation}
shows success rate, makespan, and agent-agent collisions at the largest
team size per domain: 96 agents on Random, 80 on Mazes,
192 on Warehouse, and 256 on Cities-Tiles.

Across all four domains and both training stages, most of the benefit from
additional refinement is obtained by the training depth
$K_{\mathrm{train}}=4$. A single refinement round is insufficient in the
maximum-agent settings: increasing $K_{\mathrm{test}}$ to $K_{\mathrm{test}}=4$ produces large
gains in success rate together with substantial reductions in makespan and
agent-agent collisions. Beyond four rounds, the additional improvements are
smaller. Thus, four refinement rounds capture most of the gains observed up to
$K_{\mathrm{test}}=12$, while additional rounds provide smaller improvements.

Although DMM is trained only with $K_{\mathrm{train}}=4$, the same checkpoints
remain effective when unrolled for additional refinement rounds. At
$K_{\mathrm{test}}=8$ and $12$, performance generally matches or improves on
the $K_{\mathrm{test}}=4$ result, indicating that the learned update is not
specialized to an exact four-round horizon. Changes beyond the training depth
are smaller and occasionally non-monotonic across metrics, but additional
rounds often further reduce collisions. Thus, the learned refinement dynamics
can be extended at test time without retraining.

MICPO shifts the refinement-depth trade-off toward stronger performance at
smaller $K_{\mathrm{test}}$. At the same test-time depth, DMM-MICPO-3M
generally attains higher success rates and lower makespan and collision counts
than DMM-3M. On Warehouse at 192 agents, for example,
DMM-MICPO-3M with $K_{\mathrm{test}}=2$ already achieves a lower makespan
than DMM-3M with $K_{\mathrm{test}}=12$. Additional rounds remain useful
after fine-tuning, but the MICPO checkpoint requires fewer refinement rounds
to reach a comparable level of performance.

\section{Intent-State Communication Ablation}
\label{sec:z-ablation}

\begin{figure}[H]
\centering
\includegraphics[width=\linewidth]{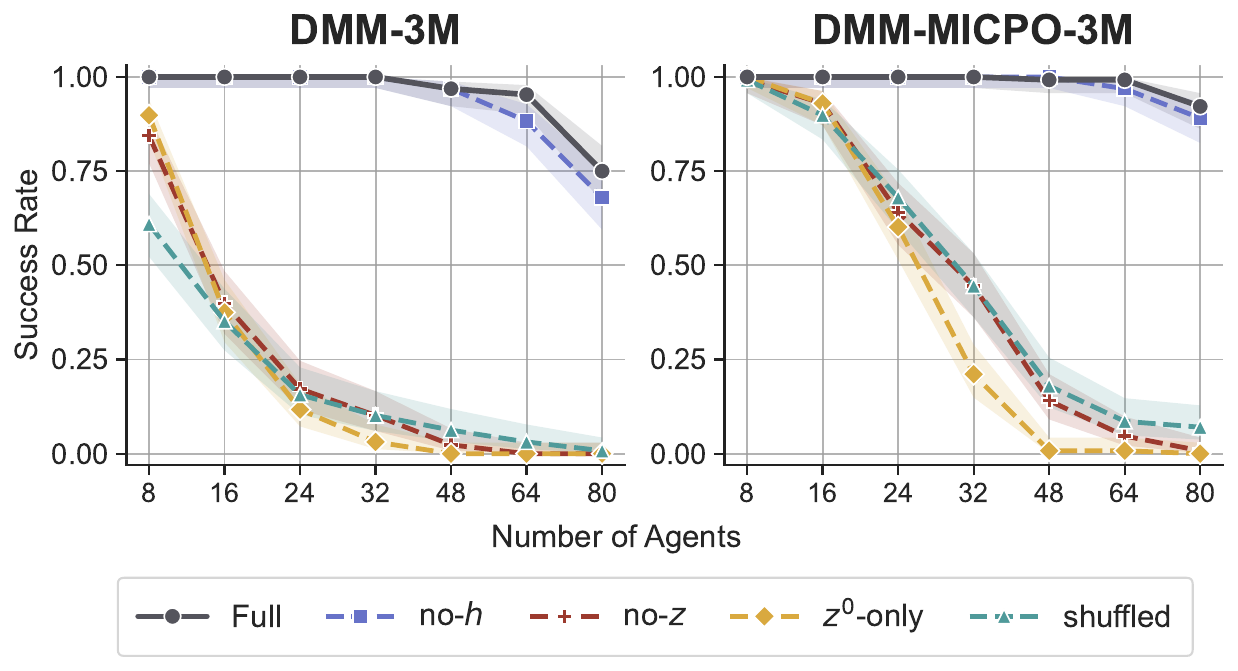}
\caption{Success rate on Mazes under inference-time perturbations of the
communicated state. Both DMM-3M and DMM-MICPO-3M are evaluated with fixed
weights at $K=4$. Full communicates both $h$ and the evolving intent $z$;
no-$h$ removes $h$ from the broadcast, no-$z$ removes $z$, and $z^0$-only
replaces the evolving communicated intent with its initial value while leaving
the agent's own intent dynamics unchanged. Shuffled permutes the assembled
messages across agents before decoding. Shaded regions show 95\% Wilson score confidence intervals across $n=128$
validation instances, with the same instances used for all compared
communication variants.}
\label{fig:z-ablation}
\end{figure}

To determine which component of the recurrent message supports coordination, we
apply inference-time perturbations to fixed DMM-3M and DMM-MICPO-3M
checkpoints without retraining, all evaluated at the trained
refinement depth $K=4$. The agent's own intent update and final action rule
remain unchanged; only the information communicated to neighboring agents is
modified. ``Full'' uses the standard message containing both
the learned feature $h$ and the projected current intent $z$; ``no-$h$''
communicates only $z$, and ``no-$z$'' communicates only $h$. In
``$z^0$-only'', each agent updates its own intent normally, but
broadcasts the fixed initial intent $z^0$ at every round instead of the
evolving $z$. Finally, ``shuffled'' permutes the assembled messages across
agents before decoding, preserving the message structure while mismatching
content with its originating agent.
Figure~\ref{fig:z-ablation} reports success rate on Mazes across team sizes.

The clearest separation is between variants that communicate the evolving
intent and those that do not: removing $h$ while retaining $z$ causes only a
limited drop relative to ``Full'', whereas ``no-$z$''
deteriorates sharply as team size grows. For these trained
policies, the evolving intent therefore carries more of the
coordination-relevant information than $h$ alone.

The ``$z^0$-only'' intervention shows this dependence is not explained
simply by providing an intent-valued message: each agent updates
its own $z$ normally, but neighbors repeatedly receive only the initial
$z^0$, and performance then degrades to a level similar to ``no-$z$''. The
relevant information thus lies in the evolution of $z$
across refinement rounds, not just its initialization. Together
with \appref{sec:round-ablation}, this supports repeated refinement
benefiting from agents communicating updated action intents across rounds.

The ``shuffled'' intervention provides a complementary test: the full
messages are preserved, but the assembled message tensors are permuted across
agents, breaking the correspondence between an agent and its constructed
message set. The resulting drop in success shows retaining
message content alone is insufficient once this correspondence is disrupted,
so effective use of the evolving intent also depends on communicating it in
the correct local context.

MICPO improves robustness to all these perturbations but does not change
their qualitative ordering: ``Full'' and ``no-$h$'' remain
stronger at large team sizes after fine-tuning, while ``no-$z$'',
``$z^0$-only'', and ``shuffled'' still deteriorate sharply. The dependence on
evolving intent communication thus persists after policy optimization,
even as MICPO improves overall performance.

These are post-training interventions on fixed checkpoints, so they
measure which communication components the learned policies rely on, not
the performance of architectures retrained without $h$, without
$z$, or with a different message parameterization.

\section{Corridor Refinement and Teacher-Forcing Ablations}
\label{sec:corridor-ablation}

\begin{figure*}[!t]
\centering
\includegraphics[width=\textwidth]{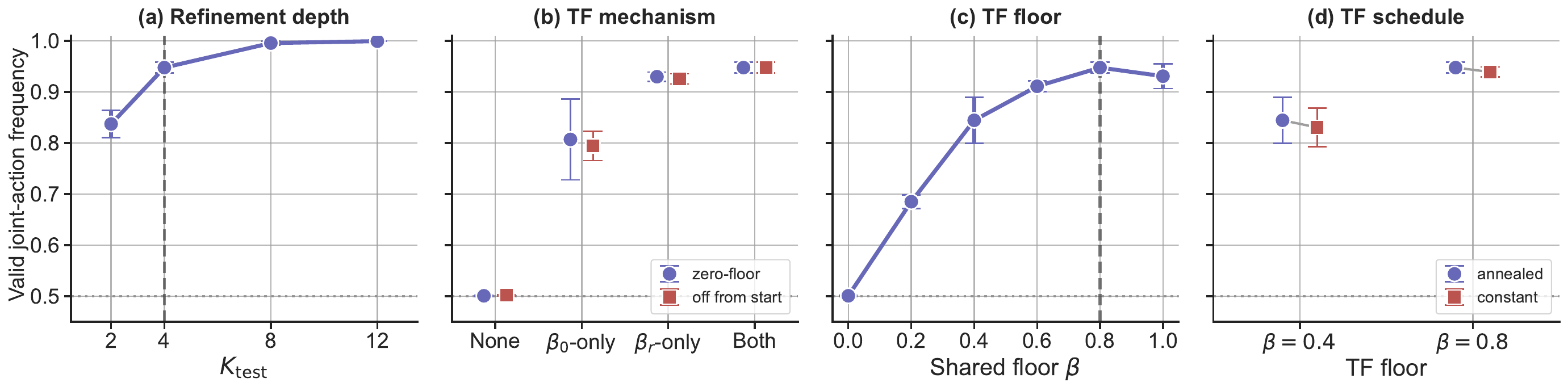}
\caption{Corridor ablations of refinement depth and teacher forcing.
All panels report the valid joint-action frequency
$p_{\mathrm{valid}}=p(\mathrm{WL})+p(\mathrm{RW})$.
(a) Checkpoints trained with $K_{\mathrm{train}}=4$, evaluated at different
test-time refinement depths without retraining; the vertical dashed line marks
the training depth.
(b) Contribution of initial-intent teacher forcing $\beta_0$ and
round-level teacher forcing $\beta_r$: for the blue series, excluded
mechanisms are annealed to a zero floor; for the red series,
they are disabled from the start of training.
(c) Effect of the shared teacher-forcing floor
$\beta_0=\beta_r=\beta$; the vertical dashed line marks the retained floor used in the main DMM
training, $\beta=0.8$.
(d) Corridor annealed schedule versus a constant
teacher-forcing probability at matched floors.
The horizontal dotted line marks the $0.5$ valid-action frequency from
independent recombination of the expert marginals.}
\label{fig:corridor-ablation}
\end{figure*}

We use the ambiguous corridor state to isolate how refinement depth and
teacher forcing affect the learned joint-action distribution
(Figure~\ref{fig:corridor-ablation}), reporting the
valid joint-action frequency
$p_{\mathrm{valid}}=p(\mathrm{WL})+p(\mathrm{RW})$,
where WL and RW are the two coordinated resolutions. Each configuration is
evaluated with 1,000 joint-action samples per training seed; point
values are means across five seeds, with error bars denoting 95\% Student-$t$
confidence intervals across seeds. The horizontal reference at
$p_{\mathrm{valid}}=0.5$ is the independent-sampling outcome for the balanced
expert marginals.

\paragraph{Refinement depth.}
With the training depth fixed at $K_{\mathrm{train}}=4$, increasing only the
test-time refinement rounds raises the valid joint-action frequency
from $83.7\pm2.7\%$ at $K_{\mathrm{test}}=2$ to
$94.8\pm1.1\%$ at $K_{\mathrm{test}}=4$, and further to
$99.6\pm0.4\%$ at $K_{\mathrm{test}}=8$ and
$99.9\pm0.1\%$ at $K_{\mathrm{test}}=12$.
The same fixed checkpoints thus continue to suppress invalid joint actions
when unrolled beyond the training depth. Table~\ref{tab:corridor-depth-composition}
shows this increase does not come from collapsing onto a single valid
resolution: both WL and RW retain substantial frequency, while the stall and
collision frequency drops from $0.16$ at $K_{\mathrm{test}}=2$ to below $0.01$
at $K_{\mathrm{test}}=12$.

\begin{table}[ht]
\centering
\caption{Effect of test-time refinement depth in the corridor, as
percentages. All checkpoints are trained with $K_{\mathrm{train}}=4$.
Entries are mean frequencies across five independently trained checkpoints
(one per training seed), with 1,000 joint-action samples per checkpoint. The
valid joint-action frequency is
$p_{\mathrm{valid}}=p(\mathrm{WL})+p(\mathrm{RW})$; uncertainty denotes the
95\% Student-$t$ confidence interval across checkpoints. WL and RW are the two
valid joint actions, while Invalid combines the stall (WW) and collision (RL)
outcomes.}
\label{tab:corridor-depth-composition}

\small
\setlength{\tabcolsep}{5.5pt}
\renewcommand{\arraystretch}{1.15}
\newcommand{\corridorvalue}[2]{#1\,{\tiny$\pm$#2}}
\begin{tabular}{@{}ccccc@{}}
\toprule
& & \multicolumn{3}{c}{Joint-action frequency (\%)} \\
\cmidrule(lr){3-5}
$K_{\mathrm{test}}$
& $p_{\mathrm{valid}}$ (\%)
& WL
& RW
& Invalid \\
\midrule
2
& \corridorvalue{83.7}{2.7}
& 41.7 & 42.1 & \corridorvalue{16.3}{2.7} \\

4 (train)
& \corridorvalue{94.8}{1.1}
& 51.4 & 43.4 & \corridorvalue{5.2}{1.1} \\

8
& \corridorvalue{99.6}{0.4}
& 52.7 & 46.9 & \corridorvalue{0.4}{0.4} \\

12
& \corridorvalue{99.9}{0.1}
& 52.9 & 47.0 & \corridorvalue{0.1}{0.1} \\
\bottomrule
\end{tabular}
\end{table}

\paragraph{Teacher-forcing mechanism.}
Round-level teacher forcing accounts for most of the coordination effect.
Annealing the excluded mechanism to zero gives a valid joint-action
frequency of $0.930$ with only $\beta_r$ active, $0.807$ with only
$\beta_0$ active, and $0.501$ with both mechanisms removed; disabling
the excluded mechanisms from the start of training gives the same ordering,
with frequencies of $0.926$, $0.794$, and $0.502$.
The standard configuration with both mechanisms reaches
$0.948$. Since $\beta_r$ determines whether the expert action is
incorporated into the evolving intent update at each round, this is
consistent with the communicated intent trajectory being the main
teacher-forced signal supporting coordinated refinement.

\paragraph{Teacher-forcing strength and schedule.}
Increasing the retained teacher-forcing floor improves the valid joint-action
frequency up to $\beta=0.8$, where the standard configuration reaches
$0.948$; at $\beta=1.0$ the frequency decreases to $0.931$. With a shared
floor $\beta_0=\beta_r=\beta$, the frequencies are $0.501$,
$0.685$, $0.845$, $0.911$, $0.948$, and $0.931$ for $\beta=0$, $0.2$,
$0.4$, $0.6$, $0.8$, and $1.0$, respectively. Annealing rather than holding
the teacher-forcing probability constant produces only small differences at
matched floors: at $\beta=0.4$, the frequencies are $0.845$ and $0.831$,
while at $\beta=0.8$ they are $0.948$ and $0.939$, respectively. Thus, the
retained floor has a much larger effect than whether the teacher-forcing
probability is annealed or held constant.

Overall, the corridor ablations show that additional test-time refinement continues to improve coordination beyond the training depth, while round-level teacher forcing provides most of the training-time coordination benefit.

\section{Decoding and Repeat-State Escape }
\label{sec:movingai-rse-comparison}

Table~\ref{tab:movingai-rse-comparison} compares three reactive-policy
configurations on the same 1,600 MovingAI task keys: sampling without RSE,
sampling with RSE, and argmax with RSE. All use PIBT-based
shielding. Every mean sum of costs (SoC) and mean makespan uses the same 1,526
tasks solved by all nine configurations, matched by task key; solved
counts use all 1,600 tasks. For HMAGAT, sampling and argmax refer to action
selection. For DMM, they refer to the four communication rounds; the final
action is greedy in all three configurations.

\begin{table}[!htb]
\centering
\caption{MovingAI decoding and RSE configurations. Solved is out of 1,600
tasks. MS: makespan.}
\label{tab:movingai-rse-comparison}
\small
\setlength{\tabcolsep}{4pt}
\begin{tabular}{@{}llrrr}
\toprule
Method & Decoding / RSE & Solved~$\uparrow$ & SoC~$\downarrow$ & MS~$\downarrow$ \\
\midrule
\multirow{3}{*}{HMAGAT}
  & Sampling         & 1,556 & 362,741 & 506.3 \\
  & Sampling + RSE   & 1,576 & 361,858 & 481.6 \\
  & Argmax + RSE     & 1,543 & 334,302 & 534.3 \\
\midrule
\multirow{3}{*}{\shortstack[l]{DMM-MICPO-\\0.8M}}
  & Sampling         & 1,593 & 369,955 & 492.6 \\
  & Sampling + RSE   & 1,592 & 369,956 & 494.4 \\
  & Argmax + RSE     & 1,593 & 321,085 & 460.0 \\
\midrule
\multirow{3}{*}{\shortstack[l]{DMM-MICPO-\\3M}}
  & Sampling         & \cellcolor{best}1,599 & 335,999 & 468.6 \\
  & Sampling + RSE   & \cellcolor{best}1,599 & 336,001 & 468.3 \\
  & Argmax + RSE     & 1,598 & \cellcolor{best}311,768 & \cellcolor{best}452.8 \\
\bottomrule
\end{tabular}
\end{table}

Adding RSE to sampling raises HMAGAT coverage by 20 tasks and lowers its
common-cohort mean makespan from 506.3 to 481.6. For DMM, the same
contrast has little effect on coverage or quality: DMM-MICPO-0.8M changes from
1,593 to 1,592 solved tasks and DMM-MICPO-3M stays at 1,599, while their matched
mean SoCs change by less than 0.2\%. In contrast, the historical argmax-with-RSE
configurations have much lower matched SoC than sampling with RSE:
321,085 versus 369,956 for DMM-MICPO-0.8M and 311,768 versus 336,001 for DMM-MICPO-3M.
HMAGAT argmax with RSE also lowers matched SoC but solves 33 fewer tasks and
has a higher matched makespan than HMAGAT sampling with RSE.

These differences describe complete inference configurations, not a
single-factor causal effect of argmax. The historical DMM argmax-with-RSE
packages use a zero initial intent, while the sampled-round runs use a
Dirichlet initial intent; the execution implementations also differ. The
nine-configuration table separates the observed RSE-on/sampling contrast from
the argmax-with-RSE contrast without attributing the latter solely to argmax.

\end{document}